\documentclass{article}
\usepackage[table,xcdraw,dvipsnames]{xcolor} 

\definecolor{cvprblue}{RGB}{33, 102, 172}       % 经典的学术深蓝，用于强调
\definecolor{softgray}{RGB}{242, 242, 242}      % 极淡的灰色，用于表格隔行
\definecolor{alertred}{RGB}{214, 39, 40}        % 柔和的警告红
\definecolor{boxback}{RGB}{245, 248, 255}       % 极淡的蓝色背景，用于定理盒子

\usepackage{microtype}      % 自动调整字符间距，减少连字符，使灰度均匀
\usepackage{booktabs}       % 专业三线表
\usepackage{multirow}
\usepackage{colortbl}       % 表格颜色

\usepackage[labelfont={bf,color=cvprblue},font=small,labelsep=period]{caption}
\usepackage{subcaption}

\usepackage[most]{tcolorbox}
\newtcolorbox{insightbox}[2][]{%
    colback=boxback,        % 背景色
    colframe=cvprblue,      % 边框色
    fonttitle=\bfseries,    % 标题粗体
    coltitle=white,         % 标题文字白色
    title={#2},             % 标题内容
    enhanced,               % 启用高级绘图
    attach boxed title to top left={yshift=-1mm, xshift=0.1mm}, % 标题位置微调
    boxrule=0.5pt,          % 边框细一点更精致
    sharp corners=south,    % 下方直角
    arc=4mm,                % 上方圆角
    top=3mm, bottom=1mm, left=1mm, right=1mm, % 内边距
    #1
}
\newcommand{\myitem}[1]{\noindent $\bullet$ #1}

\usepackage{microtype}
\usepackage{graphicx}
\usepackage{subcaption}
\usepackage{booktabs} % for professional tables

\usepackage{hyperref}

\usepackage{pgfplots}
\pgfplotsset{compat=1.17} % 或更新版本
\usepackage{tikz}
\usetikzlibrary{shapes.geometric}

\usepackage[preprint]{icml2026}

\usepackage{amsmath}
\usepackage{amssymb}
\usepackage{mathtools}
\usepackage{amsthm}
\usepackage{colortbl}
\definecolor{highlight}{gray}{0.9}
\usepackage[capitalize,noabbrev]{cleveref}
\usepackage{multirow}
\theoremstyle{plain}
\newtheorem{theorem}{Theorem}[section]
\newtheorem{proposition}[theorem]{Proposition}
\newtheorem{lemma}[theorem]{Lemma}

\theoremstyle{definition}

\theoremstyle{remark}

\usepackage[textsize=tiny]{todonotes}

\icmltitlerunning{WaterVIB: Learning Minimal Sufficient Watermark Representations}

\begin{document}

\twocolumn[
  %\icmltitle{WaterVIB: Learning Minimal Sufficient Watermark Representations \\
  %         via Variational Information Bottleneck}
    % 第一个大括号里保留 \\ 用于正文排版，第二个大括号用空格或冒号用于PDF属性
\icmltitle{WaterVIB: Learning Minimal Sufficient Watermark Representations \texorpdfstring{\\}{: } via Variational Information Bottleneck}



  \icmlsetsymbol{equal}{*}

  \begin{icmlauthorlist}
   \icmlauthor{Haoyuan He}{tsinghua}
   \icmlauthor{Yu Zheng}{tsinghua}
   \icmlauthor{Jie Zhou}{tsinghua}
   \icmlauthor{Jiwen Lu}{tsinghua}
  \end{icmlauthorlist}
% --- 作者列表开始 ---

  \icmlaffiliation{tsinghua}{Department of Automation, Tsinghua University}

  %\icmlcorrespondingauthor{Jie Zhou}{jzhou@tsinghua.edu.cn}
  %\icmlcorrespondingauthor{Jiwen Lu}{lujiwen@tsinghua.edu.cn}

  % paper; these are used to populate the ``keywords'' metadata in the PDF but
  \icmlkeywords{Variational Information Bottleneck, Robust Watermarking, Deep Learning, ICML}

  \vskip 0.3in ]



\printAffiliationsAndNotice{}  % no special notice (required even if empty)

\begin{abstract}
Robust watermarking is critical for intellectual property protection, whereas existing methods face a severe vulnerability against regeneration-based AIGC attacks. 
% This process acts as a semantic projection that effectively cleanses the non-natural artifacts standard encoders rely on. 
We identify that existing methods fail because they entangle the watermark with high-frequency cover texture, which is susceptible to being rewritten during generative purification.  
To address this, we propose WaterVIB, a theoretically grounded framework that  reformulates the encoder as an information sieve via the Variational Information Bottleneck. 
Instead of overfitting to fragile cover details, our approach forces the model to learn a \textbf{Minimal Sufficient Statistic} of the message. 
This effectively filters out redundant cover nuances prone to generative shifts, retaining only the essential signal invariant to regeneration. 
We theoretically prove that optimizing this bottleneck is a necessary condition for robustness against distribution-shifting attacks. 
Extensive experiments demonstrate that WaterVIB significantly outperforms state-of-the-art methods, achieving superior zero-shot resilience against unknown diffusion-based editing. 

\end{abstract}

\section{Introduction} 
Digital watermarking serves as a crucial technology for copyright protection and content provenance in the era of digital media. 
Ideally, a watermark should be imperceptible to humans yet robust enough to survive various distortions. 
Deep learning has significantly advanced this field by jointly training encoder-decoder networks, thereby embedding messages into images with high fidelity~\citep{bui2023trustmark}. 

While deep learning watermarking has achieved resilience against standard distortions (e.g., Gaussian noise, JPEG compression) 
via heuristic data augmentation~\citep{zhu2018hidden},
this paradigm struggles to withstand the emerging threat of generative purification~\citep{podell2023sdxl}.  %这里还要引用对应的水印清除工作
% has introduces a new category of threat that challenges these approaches. 
Unlike traditional attacks that merely degrade image quality, modern diffusion-based tools~\citep{Zhao2023InvisibleIW} regenerate content leveraging learned priors~\citep{rombach2022high}, stripping away watermark signals while preserving visual fidelity. 
By projecting the watermarked image back onto the manifold of natural images, these methods eliminate hidden signals perceived as unnatural perturbations.
%Consequently, existing models that overfit to known noise patterns fail to generalize to these complex semantic modifications, leading to the erasure of copyright information.
Consequently, existing watermarks, trained primarily on additive or geometric distortions, may well fail to generalize to such semantic regeneration, resulting in the inevitable erasure of copyright information~\citep{Zhao2023InvisibleIW}.

\begin{figure}[t]  % 推荐放在浮动体环境中
    \centering
    \includegraphics[width=0.5\textwidth]{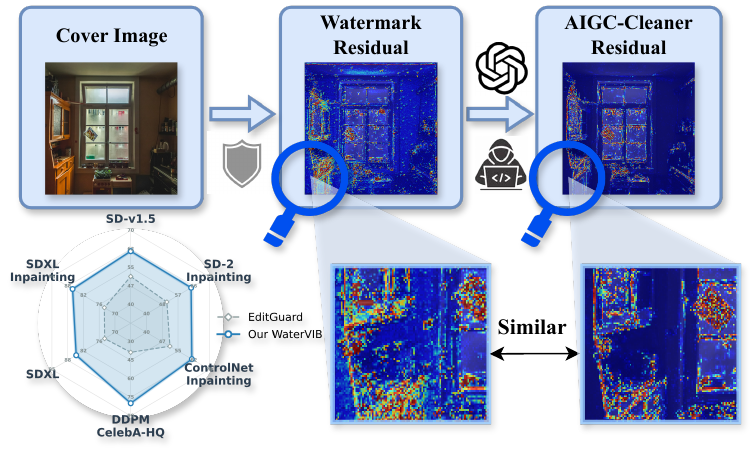}
    \caption{\textbf{Vulnerability of standard watermarking versus the robustness of our VIB method.} The residual visualizations (Top) reveal a strong correlation between the watermark signal and AIGC purification, highlighting the fragility of texture-entangled methods. We visualize the Bit Accuracy (higher is better) across six AIGC editing benchmarks. To clearly demonstrate the relative improvements on tasks with varying difficulty levels, each axis is independently normalized to its effective range}
    \label{fig1}
\end{figure}

We attribute this vulnerability to a structural flaw in how standard encoders are trained. 
Conventional end-to-end schemes typically minimize a pixel-wise reconstruction error combined with a message decoding loss.
To satisfy the invisibility constraint, the encoder inadvertently learns to hide the watermark signal within the high-frequency textures of the cover image~\citep{zhang2020udh}, as the human eye is less sensitive to changes in these complex regions. 
However, this strategy creates a fatal dependency, i.e., the message becomes entangled with specific local details.
As shown in Figure~\ref{fig1}, generative models act as 'manifold projectors' that specifically synthesize and rewrite these texture regions to improve perceptual quality~\citep{nie2022diffusion}.  
When the cover image's texture is regenerated, the spurious correlations between the message and the cover details are severed, thus destroying the watermark.  

To address this challenge, we propose to explicitly disentangle the watermark from the fragile cover content. 
Rather than hiding signals in high-frequency textures prone to regeneration, the encoder should anchor the message to the robust semantic structure of the image. 
Formally, we frame this as learning a \textbf{Minimal Sufficient Statistic (MSS)} of the message with respect to the cover. 
This objective seeks a representation that retains minimal information about the cover while remaining sufficient for decoding, which we theoretically prove is equivalent to optimizing the \textbf{Information Bottleneck (IB)} principle~\citep{tishby2000information}. 
% We model this optimization problem from an information-theoretic perspective and derive that it is equivalent to optimizing the Information Bottleneck (IB) objective. 
Guided by this insight, we introduce WaterVIB, a principled framework that reformulates the watermarking encoder as an information sieve. 
By minimizing a variational upper bound, WaterVIB forces the model to filter out redundant cover nuances susceptible to generative shifts. 
This ensures that the model retains only the essential watermark signal that is robust to generative purification. 
Consequently, this theoretically grounded design enables WaterVIB to achieve superior zero-shot resilience against unknown generative purification, significantly outperforming state-of-the-art methods. 

\textbf{Our contributions are summarized as follows:}

\myitem{We identify the texture entanglement phenomenon as a major cause of failure for existing watermarking methods against generative editing, and introduce WaterVIB framework that leverages the Information Bottleneck principle to learn a robust, disentangled watermark representation. }

\myitem{We provide a theoretical analysis showing that the learning objective of our WaterVIB framework acts as a necessary condition for robustness against generative variations within our framework. }

\myitem{Extensive experiments demonstrate that WaterVIB significantly outperforms SOTA baselines, showing superior resilience against both known distortions and unknown generative purification, without specific adversarial training.}

\section{Related Works}
\subsection{Deep Image Watermarking}
Traditional watermarking schemes typically embedded signals into specific frequency domains, such as DCT~\citep{langelaar2001optimal} , DWT~\citep{ganic2004robust} , or DFT~\citep{kang2003dwt}, to balance imperceptibility and robustness. The advent of deep learning revolutionized this field by introducing end-to-end optimization pipelines~\citep{bui2023trustmark}. Pioneer works like HiDDeN~\citep{zhu2018hidden} and StegaStamp~\citep{tancik2020stegastamp}  utilized encoder-decoder architectures trained with differentiable noise layers. Building on this, recent approaches have integrated GANs and Invertible Neural Networks~\citep{jing2021hinet}  to enhance visual quality and recovery accuracy. To address advanced security threats, state-of-the-art methods have expanded into specialized domains: EditGuard~\citep{zhang2024editguard} introduces dual watermarks for tamper localization, while NeRF-Signature~\citep{luo2025nerf}  extends embedding capabilities to 3D neural radiance fields. 

Despite these architectural advancements, most existing methods rely on empirical strategies to train networks against noise without explicitly decoupling the information entanglement between the cover image and the watermark. Consequently, the embedded signals often remain statistically dependent on the cover image redundancies~\citep{zhang2020udh}, leading to vulnerability against noise from distribution shifts in specific frequency domains.

\subsection{Information Bottleneck Principle}
The Information Bottleneck (IB) principle, originally introduced by~\citep{tishby2000information}, formulates representation learning as an optimization trade-off. This theoretical framework was later extended to deep neural networks by \citep{alemi2016deep}, who proposed the Variational Information Bottleneck (VIB). By leveraging variational inference and the reparameterization trick, VIB made the estimation of mutual information bounds computationally tractable for high-dimensional data. Since then, the VIB framework has been successfully applied to diverse domains, ranging from model compression \citep{razeghi2022compressed} and interpretability~\citep{seo2023interpretable}  to temporal video analysis~\citep{zhong2023semantic} and large language model alignment~\citep{ yang2025exploring}. Unlike these predominant classification or recognition tasks where the target is a label, we adapt the IB framework to the watermarking channel to construct a signal explicitly decoupled from cover image redundancies. To our knowledge, WaterVIB is the first to rigorously bridge Information-Theoretic Representation Learning and deep generative watermarking.

\section{Theoretical Analysis}
\label{sec:theory}
In this section, we formally characterize the vulnerability of existing watermarking methods under generative purification. 
We posit that the failure of current methods stems from a \textit{distributional entanglement} with the cover image's texture. 
Consequently, we propose to resolve this conflict by enforcing the Information Bottleneck principle, aiming to approximate the Minimal Sufficient Statistic (MSS) of the message. 

\begin{table}[ht]
    \centering
    \caption{\textbf{Evidence of Signal-Spatial Alignment.} Top: Spectral energy distribution shows the AIGC distortion ($s_{atk}$) targets the same high-frequency bands as the watermark ($\delta$). Bottom: Pearson Correlation (PCC) shows the attack is structurally dependent on the cover image, unlike random noise.}
    \label{tab:alignment_evidence}
    \vspace{2mm}
    \resizebox{0.48\textwidth}{!}{
    \begin{tabular}{lcccc}
        \toprule
        \multicolumn{5}{c}{\textbf{(a) Spectral Energy Distribution Alignment}} \\
        \midrule
        \textbf{Signal Component} & \textbf{High Freq.} & \textbf{Mid Freq.} & \textbf{Low Freq.} & \textbf{Dominant} \\
        \midrule
        \rowcolor{highlight}
        Watermark Signal ($\delta$) & 30.20\% & \textbf{45.45\%} & 24.35\% & \textbf{Mid-High} \\
        \rowcolor{highlight}
        AIGC Distortion ($s_{atk}$) & 23.36\% & 20.49\% & \textbf{56.15\%} & \textbf{Mid-High*} \\
        Natural Images ($x$) & 1.16\% & 11.92\% & \textbf{86.91\%} & Low \\
        \bottomrule
        \\[-2mm] % Add some vertical space
        \toprule
        \multicolumn{5}{c}{\textbf{(b) Structural Dependence (PCC with Cover Image)}} \\
        \midrule
        \textbf{Target Signal} & \multicolumn{3}{c}{\textbf{Pearson Correlation Coefficient}} & \textbf{Interpretation} \\
        \midrule
        \rowcolor{highlight}
        AIGC Distortion ($s_{atk}$) & \multicolumn{3}{c}{\textbf{0.5989}} & \textbf{Content-Dependent} \\
        Watermark Signal ($\delta$) & \multicolumn{3}{c}{0.4535} & Entangled \\
        Random Noise ($n$) & \multicolumn{3}{c}{-0.0003} & Independent \\
        \bottomrule
    \end{tabular}
    }
    \vspace{-3mm}
\end{table}

\subsection{Analysis of Vulnerability}
\label{subsec:theory-1}
Let $\mathcal{X} \subseteq \mathbb{R}^{H \times W \times C}$ be the image space and $\mathcal{M}$ be the message space. 
% Assuming natural images lie on a low-dimensional manifold $\mathcal{M} \subset \mathcal{X}$, 
%Consistent with the spectral bias of neural networks,
A standard encoder $E: \mathcal{X} \times \mathcal{M} \to \mathcal{X}$ produces a watermarked image $x_{wm} = x + \delta$. 
Unlike traditional additive noise, we model Generative Purification as a projection operator onto the natural image space. 
Let $p_{data}(x)$ be the natural image distribution. The purification process $G: \mathcal{X} \to \mathcal{X}$ projects $x_{wm}$ onto the high-density region of $p_{data}$ by minimizing a perceptual distance $d(\cdot, \cdot)$: 
\begin{equation}
\label{eq:purification_constraint}
    x_{rec} = \mathcal{G}(x_{wm}) \approx \arg\min_{x' \in \text{supp}(p_{data})} d(x', x_{wm}).
\end{equation}

Current methods implicitly assume the watermark signal $\delta$ resides in a subspace orthogonal to the semantic content. However, we propose that generative models preferentially reconstruct low-frequency semantics while rewriting high-frequency textures where $\delta$ resides. 

\newtheorem{observation}{Key Observation} % 用于替代伪 Proposition

\begin{insightbox}[label=prop:alignment]{Observation: Spectral and Spatial Alignment}
Let $\mathbf{s}_{atk} = \mathcal{G}(\mathbf{x}_{wm}) - \mathbf{x}_{wm}$ be the purification distortion. 
We observe that watermark failure is driven by a \textbf{Spectral and Spatial Alignment}, where both the watermark signal $\delta$ and attack distortion $s_{atk}$ reside in the intersection of the high-frequency subspace $\mathcal{V}_H$ and the texture-dependent subspace.
\end{insightbox}
\vspace{-0.5em}
\iffalse
\begin{proposition}
    [\textbf{Spectral and Spatial Alignment}]
\label{prop:alignment}
Let $\mathbf{s}_{atk} = \mathcal{G}(\mathbf{x}_{wm}) - \mathbf{x}_{wm}$ be the purification distortion. 
We observe that watermark failure is driven by a Spectral and Spatial Alignment, where both the watermark signal $\delta$ and attack distortion $s_{atk}$ reside in the intersection of the high-frequency subspace $\mathcal{V}_H$ and the texture-dependent subspace.  
% The perturbation introduced by generative purification, denoted as $s_{atk} = \mathcal{G}(x_{wm}) - x_{wm}$, is non-isotropic. Specifically, $s_{atk}$ exhibits high statistical correlation with the watermark signal $\delta$ in both the frequency domain and the local texture space.
\end{proposition}
\fi

\textit{Empirical Validation.} We validate Observation~\ref{prop:alignment} via spectral analysis and Pearson Correlation Coefficient (PCC). 
As reported in \textbf{Table~\ref{tab:alignment_evidence}(a)}, the watermark signal(EditGuard, \citet{zhang2024editguard}) concentrates $75.65\%$ of its energy in Mid-High frequencies. 
Meanwhile, the AIGC distortion $s_{atk}$ effectively targets these same bands ($43.85\%$ energy), unlike natural images.
Furthermore, \textbf{Table~\ref{tab:alignment_evidence}(b)} shows a high spatial correlation between the distortion and the cover image ($\rho\approx0.60$), implying that the attack is content-dependent. 
\subsection{Gradient Counter-Optimization}
\label{subsec:collapse}
The alignment described in Proposition~\ref{prop:alignment} leads to a direct interference with the decoding objective. Let $\mathcal{L}(\cdot)$ be the decoding loss, which is the BCE loss between the target message $\mathbf{m}$ and the extracted message~\citep{sander2025watermark}.

\begin{insightbox}[label=lemma:gradient]{Lemma 3.2: Gradient Interference}
For a watermarked image $x_{wm}$ optimized to minimize $\mathcal{L}$, the robustness loss under generative attack is approximated by the inner product $\langle s_{atk}, \nabla_x \mathcal{L}(x_{wm}) \rangle$. A positive inner product implies \textit{counter-optimization}.
\end{insightbox}
\vspace{-0.5em}
\iffalse
\begin{lemma}[\textbf{Gradient Interference}]
\label{lemma:gradient}
For a watermarked image $x_{wm}$ optimized to minimize $\mathcal{L}$, the robustness loss under generative attack is approximated by the inner product $\langle s_{atk}, \nabla_x \mathcal{L}(x_{wm}) \rangle$. A positive inner product implies \textit{counter-optimization}.
\end{lemma}
\fi

% This robustness collapse can be further quantified and theoretically justified. We demonstrate this by analyzing the sensitivity of the decoder $\mathcal{D}: \mathcal{X} \to \mathbb{R}^L$ to perturbations in $\mathcal{S}_{overlap}$. Let the decoding objective to be the BCE loss $\mathcal{L}(\cdot)$ between the target message $\mathbf{m}$ and the extracted message.
\noindent\textit{Proof.} 
Consider the watermarked image $\mathbf{x}_{wm} = \mathbf{x}_{cover} + \mathbf{s}_{wm}$, and the AIGC purified image $\mathbf{x}_{atk} = \mathbf{x}_{wm} + \mathbf{s}_{atk}$. 
Assuming $\mathcal{L}$ is differentiable, we apply a first-order Taylor expansion to the loss $\mathcal{L}(\mathbf{x}_{atk})$ around $\mathbf{x}_{wm}$:
\begin{equation}
\label{eq:1-order expansion}
    \mathcal{L}(\mathbf{x}_{atk}) \approx \mathcal{L}(\mathbf{x}_{wm}) + \mathbf{s}_{atk}^\top \nabla_{\mathbf{x}} \mathcal{L}(\mathbf{x}_{wm})
\end{equation}
Since the encoder minimizes $\mathcal{L}$, the gradient $\nabla_x \mathcal{L}$ points in the direction of increasing decoding error. 
Consequently, if the projection of the attack vector $s_{atk}$ onto the gradient is positive, the loss increases and signifies a successful attack. 

\textit{Empirical Validation.}
We quantify this interference through the scalar projections of both the watermark and AIGC distortion onto the gradient direction in \textbf{Table~\ref{tab:gradient_interference_1}}. 
While the watermark signal $\delta$ naturally has a negative projection ($-0.063$) to minimize loss, the AIGC distortion $s_{atk}$ shows a significant \textbf{positive projection} ($+0.027$). 
This confirms Lemma~\ref{lemma:gradient} where generative purification acts as an adversarial attack, canceling out $\approx 42.9\%$ of the optimization effort. 

\iffalse
Empirically, we calculate the average cosine similarity as  $\mathbb{E}\big[\cos(\mathbf{s}_{wm}, \nabla_{\mathbf{x}}\mathcal{L}(\mathbf{x}_{wm}))\big] \approx -0.42$. This substantial alignment confirms that the decoder $\mathcal{D}(\cdot)$ is primarily sensitive to variations along the watermark direction.

However, the magnitude of the attack reveals a critical disparity. As shown in Table \ref{tab:gradient_projection}, we quantify the scalar projections of both the watermark and AIGC distortion onto the gradient direction. The AIGC signal introduces a significant \textit{positive} projection ($+0.027$) that directly opposes the watermark's \textit{negative} projection ($-0.063$). This results in a high interference ratio of $\gamma \approx 42.9\%$. By substituting this into Eq. \ref{eq:1-order expansion}, it becomes evident that the AIGC noise counteracts nearly half of the watermark's optimization effort, precipitating the observed robustness collapse.
\fi
 %这一段和表格后的一段应该合并，我将其改成As shown in table 3 balbala，
\begin{table}[t]
    \centering
    \caption{\textbf{Gradient Counter-Optimization Analysis.} We project the signals onto the loss gradient direction $\nabla_x \mathcal{L}$. A negative projection implies loss minimization, while a \textbf{positive} projection implies loss maximization.}
    \label{tab:gradient_interference_1}
    \vspace{2mm}
    \resizebox{0.48\textwidth}{!}{
    \begin{tabular}{lccc}
        \toprule
        \textbf{Vector Component} & \textbf{Projection Value} & \textbf{Direction} & \textbf{Effect on Decoding} \\
        \midrule
        Gradient Direction ($\nabla_x \mathcal{L}$) & N/A & Target & N/A \\
        \midrule
        Watermark Signal ($\delta$) & $-0.063$ & Anti-Gradient & \textcolor{blue}{Minimizes Error} \\
        \rowcolor{highlight}
        \textbf{AIGC Distortion ($s_{atk}$)} & $\mathbf{+0.027}$ & \textbf{Gradient} & \textcolor{red}{\textbf{Maximizes Error}} \\
        \bottomrule
    \end{tabular}
    }
\end{table}

\iffalse
\begin{table}[ht]
    \centering
    \caption{Mean Scalar Projections onto the Gradient. A significant positive projection for AIGC indicates a strong adversarial effect.}
    \label{tab:gradient_projection}
    \begin{tabular}{l c c}
        \toprule
        \textbf{Projection Length} & \textbf{Watermark} & \textbf{AIGC} \\
        \midrule
        Projection on $\nabla_{\mathbf{x}}\mathcal{L}$ & -0.063 & 0.027 \\
        \bottomrule
    \end{tabular}
\end{table}
\fi
%The AIGC signal severely interferes with the watermark, yielding a high noise-to-signal ratio along the decoding gradient: $\gamma = 0.027/0.063 \approx 42.9\%$. Substituting this into Eq.~\ref{eq:1-order expansion}, the AIGC-induced loss increase effectively cancels out nearly half ($\gamma$) of the watermark's contribution to the decoding objective, precipitating the observed robustness collapse.

%有两个事实值得放在附录，一个是AIGC和watermark有一个负的cos夹角，另一个是AIGC和原图之间逐像素的PCC是负的，意味着AIGC试图抹除原图的边缘细节，这也导致了水印的失效

\subsection{Minimal Sufficient Statistic}
\label{subsec:mss_ib}
The analysis in Section~\ref{subsec:collapse} reveals that the vulnerability of current watermarks stems from their dependency on the high-frequency textures of the cover image $X$. 
Since generative purification $G(\cdot)$ acts as a manifold projection that rewrites these textures based on the prior $p_{data}(x)$, any watermark signal entangled with $X$ is inevitably corrupted. 
To defend against this, we seek a representation $Z$ that is \textit{disentangled} from the cover $\mathbf{x}$ details while remaining \textit{predictive} of the message $M$. 
Specifically, we formulate the watermark extraction process not merely as signal recovery, but as extracting the \textbf{Minimal Sufficient Statistic (MSS)} of the message from the watermarked content. 
%Formally, we define this optimal $Z$ as the \textbf{Minimal Sufficient Statistic (MSS)} of $M$ given $X$, which must satisfy two information-theoretic properties:

\begingroup
\setlength{\abovedisplayskip}{4pt}
\setlength{\belowdisplayskip}{4pt}
\vspace{0.3em}
\noindent \textbf{(1) Sufficiency (Robustness).} To ensure feasible decoding, $Z$ must capture all information necessary to decode $M$:
\begin{equation}
\label{eq:sufficiency}
    I(Z; M) = I(X; M). 
\end{equation}

\noindent \textbf{(2) Minimality (Disentanglement).} To evade texture-biased purification, $Z$ must strip away nuisance cover textures by minimizing information about $X$:
\begin{equation}
\label{eq:minimality}
    I(Z; X) \le I(\tilde{Z}; X), \; \forall \tilde{Z} \text{ s.t. } I(\tilde{Z}; M) = I(X; M). 
\end{equation}
\endgroup

\noindent The theoretical justification for imposing these constraints is established by the following theorem.    

\begin{insightbox}[label=thm:mss_optimality]{Theorem 3.3: Optimality of MSS Representation}
Let the image domain $\mathcal{X}$ be discrete and the decoding function $D(\cdot)$ be deterministic. A representation $Z$ that satisfies the Sufficiency (Eq.~\ref{eq:sufficiency}) and the Minimality (Eq.~\ref{eq:minimality}) condition is precisely the Minimal Sufficient Statistic (MSS) of $X$ for $M$.
\end{insightbox}
\vspace{-0.5em}
\noindent\textit{Proof.}
    We provide the rigorous derivation in Appendix~\ref{sec:appendix_proof}. 
    %This condition enforces maximal disentanglement, stripping away the high-frequency cover details that are prone to generative purification.

However, direct computation of the MSS is intractable in high-dimensional spaces. We therefore relax the strict sufficiency constraint to an $\epsilon$-approximate version:
\begin{equation}
\label{eq:constrained_optimization}
\begin{aligned}
    \min_{p(z|x)} \quad & I(Z; X) \\
    \text{s.t.} \quad & I(Z; M) \ge I(X; M) - \epsilon.
\end{aligned}
\end{equation}

Leveraging the convexity of mutual information (see Appendix~\ref{sec:appendix_1_proof}), we apply Lagrange multipliers to transform this into the unconstrained Information Bottleneck (IB) objective with $\beta(\epsilon) > 0$: 

\begin{equation}
\label{eq:ib_objective}
    \max_{p(z|x)} \mathcal{L}_{IB} = I(Z; M) - \beta I(Z; X)
\end{equation}

Here, maximizing $I(Z; M)$ ensures the robustness of the watermark (\textbf{sufficiency}), while minimizing $I(Z; X)$ enforces disentanglement (\textbf{minimality}), explicitly filtering out the texture of the cover prone to generative purification.

\section{Approach}
\label{sec:approach}

\begin{figure*}[t]
    \centering
    % width=\textwidth 让图片宽度等于整个页面的文字宽度
    % 如果觉得太大，可以改成 0.9\textwidth 或 0.95\textwidth
    \includegraphics[width=\textwidth]{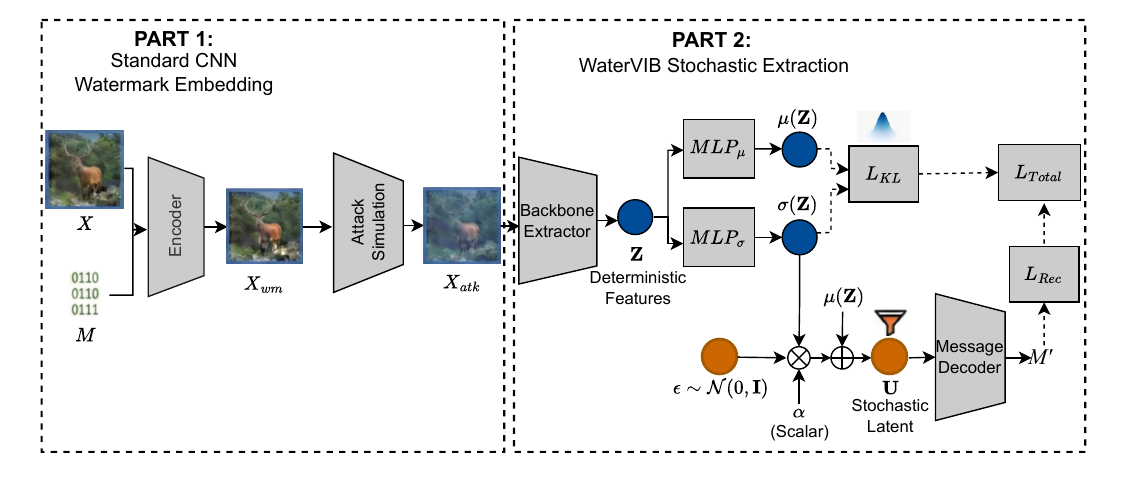} 
    
    % 可选：如果你觉得图片下方的标题离图片太远，可以用负的vspace微调
    % \vspace{-0.2cm} 
    
    \caption{\textbf{The WaterVIB Architecture.} We propose a Stochastic Information Sieve mechanism (Part 2) to defend against generative purification (Part 1). By injecting noise via a learnable bottleneck layer, WaterVIB penalizes the retention of cover-specific details ($I(Z;X)$) via the Information Bottleneck principle. This explicitly disentangles the watermark signal from the cover texture, yielding a stochastic representation $\mathbf{U}$ that is invariant to the semantic projections performed by diffusion models.}
    \label{fig:pipeline}
    
    % 可选：如果你觉得标题离下文太近或太远，可以在这里微调
    % \vspace{-0.3cm}
\end{figure*}

We operationalize the theoretical MSS objective by instantiating the \textbf{WaterVIB} framework. Unlike standard deterministic encoders, we introduce a novel stochastic bottleneck layer that functions as a differentiable \textit{information sieve}, explicitly filtering out fragile redundancies via the variational bounds derived below.

\subsection{Variational Lower Bounds}
Direct computation of the mutual information terms in $\mathcal{L}_{IB}$ \ref{eq:ib_objective} is intractable. We therefore employ variational inference to derive differentiable bounds for both components.

\textbf{Maximizing Relevance ($\mathcal{L}_{rec}$).}
Maximizing $I(Z; M)$ minimizes the conditional entropy $H(M|Z)$. We parameterize the posterior directly via the decoder. Treating the message as a vector $\mathbf{m} \in \{0,1\}^L$ and the prediction as $\hat{\mathbf{m}}$, we derive the Binary Cross-Entropy (BCE) loss:

\begin{equation}
\label{eq:L_rec}
\begin{aligned}
    \mathcal{L}_{rec} &= \mathbb{E}_{p(z, \mathbf{m})} \left[ - \log p(\mathbf{m}|z) \right] \\
    &= \mathbb{E} \left[ - \mathbf{m}^\top \log \hat{\mathbf{m}} - (\mathbf{1}-\mathbf{m})^\top \log (\mathbf{1}-\hat{\mathbf{m}}) \right]
\end{aligned}
\end{equation}

\textbf{Minimizing Compression ($\mathcal{L}_{KL}$).}
We minimize the compression term $I(Z; X)$ by optimizing its variational upper bound derived from the VIB framework~\cite{alemi2016deep}. Specifically, we introduce a fixed prior $r(z)$ (e.g., $\mathcal{N}(0, I)$) to approximate the intractable marginal $p(z)$:

\begin{align}
    I(Z; X) &= \mathbb{E}_{x} \left[ D_{KL}(p_\theta(z|x) \| p(z)) \right] \nonumber \\
    &= \mathbb{E}_{x} \left[ \int p_\theta(z|x) \log \frac{p_\theta(z|x)}{r(z)} dz \right] - \underbrace{D_{KL}(p(z) \| r(z))}_{\ge 0} \nonumber \\
    &\le \mathbb{E}_{x} \left[ D_{KL}(p_\theta(z|x) \| r(z)) \right] \triangleq \mathcal{L}_{KL} \label{eq:vib_bound}
    \raisetag{15pt} % 如果还是换行，可以调整这个数值（如 20pt）
\end{align}

The final training objective combines these terms: 
\begin{equation}
    \mathcal{L}_{total} = \mathcal{L}_{rec} + \beta \mathcal{L}_{KL}
\end{equation}

\subsection{Stochastic Bottleneck Implementation}

To implement the probabilistic encoder $p_\theta(z|x)$ within a deterministic deep neural network, we construct a \textbf{stochastic layer} $U(Z)$ (Figure~\ref{fig:pipeline}), where $Z = E_{det}(X)$ denote the deterministic features extracted by the backbone encoder.

Standard backpropagation cannot flow through random sampling. To resolve this, we employ the reparameterization trick \cite{kingma2013auto}. Furthermore, to control the variance magnitude and stabilize the adversarial training process involving watermarks, we introduce a scaling factor $\alpha$. The latent variable $U$ is sampled as:
\begin{equation} 
\label{eq:sample}
    U = \mu(Z) + \alpha \cdot \epsilon \odot \sigma(Z), \quad \epsilon \sim \mathcal{N}(0, I)
\end{equation}
where $\epsilon$ is sampled from a standard normal distribution.
This formulation allows gradients to propagate deterministically through $\mu$ and $\sigma$ while isolating the stochasticity in $\epsilon$. 

Specifically, we derive the distributional parameters $\mu(Z)$ and $\sigma(Z)$ by projecting the deterministic features $Z$ through two parallel \textbf{Multi-Layer Perceptrons (MLPs)}:
\begin{equation}
    \mu(Z) = \text{MLP}_{\mu}(Z), \quad \sigma(Z) = \exp\left(\frac{1}{2}\text{MLP}_{\sigma}(Z)\right)
\end{equation}
The reconstruction head then decodes the message $M'$ solely from the compressed stochastic representation $U$, forcing the network to retain only the robust, minimal sufficient statistics of the watermark.

\textbf{Training vs. Inference.} 
To enforce the IB constraint, we employ the stochastic sampling mechanism in \eqref{eq:sample} during training, which encourages the filtration of redundant entangled signals. At inference time, we replace this with a deterministic mapping $U = \mu(Z)$ to prevent stochastic fluctuations.

\section{Evaluation}

\subsection{Experimental Setup}

\textbf{Datasets and Models.}  Following \cite{zhu2018hidden,zhang2024editguard}, we utilize \textbf{COCO}~\citep{lin2014microsoft} for training. For evaluation, we employ both the standard COCO test set and the \textbf{AGE-Set}~\citep{zhang2024editguard} to strictly assess zero-shot robustness against AIGC manipulation. To demonstrate universality, we integrate WaterVIB into two representative backbones: \textbf{HiDDeN}(lightweight)~\citep{zhu2018hidden} and \textbf{EditGuard}~\citep{zhang2024editguard} (high-capacity SOTA). We benchmark performance against methods including TrustMark~\citep{bui2023trustmark}, WM-A~\citep{sander2025watermark}, DWT-DCT-SVD~\citep{kang2003dwt}, and so on.

\textbf{Attacks and Metrics.}  We evaluate robustness against two categories of distortions: (1) Standard Noises (e.g., JPEG, Crop, Resize, Dropout)\citep{zhu2018hidden} , and (2) Generative Purification, simulated via \textbf{diffusion}\cite{nie2022diffusion} pipelines (e.g., SD-Inpaint, DiffPure)  to effectively rewrite image content. We report \textbf{Bit Error Rate} (BER) to measure watermark survival, \textbf{PSNR} and \textbf{SSIM} to quantify visual imperceptibility.

%\textbf{VIB Settings.}   We implement the WaterVIB module as a plug-and-play stochastic layer, tailored to the specific architecture of the backbone network.

%\textbf{High-Capacity Architecture (EditGuard):}To handle the high-dimensional feature maps, we insert the VIB module after the 16-channel output of the bit decoder using a CNN-based compression pipeline. The input features ($16\times400\times400$) are first downsampled to $128\times128$ and compressed via channel reduction layers ($Conv_{16\rightarrow4} \rightarrow Conv_{4\rightarrow2}$). These condensed features are then mapped to distributional parameters $\mu$ and $\log\sigma^2$ to sample the stochastic latent $U$. Finally, the spatial structure is flattened and projected via a linear layer to the 100-bit watermark output. We set the parameters to $\alpha=10^{-4}$ and $\beta = 0.0003$.

%\textbf{Lightweight Architecture (HiDDeN):} For the lightweight HiDDeN backbone, the VIB module is integrated between the feature extractor and the final readout layer. We set the dimension of the stochastic latent variable to 128. To balance information retention and compression, we apply the parameters of $\alpha=0.007$ and $\beta=0.00015$ during the reparameterization step.

\textbf{VIB Settings.} We implement the WaterVIB module as a plug-and-play stochastic layer tailored to each backbone. 

For the high-capacity \textbf{EditGuard}, we insert a CNN-based VIB module after the 16-channel bit decoder output. To handle high-dimensional features, we employ a channel-reduction pipeline ($Conv_{16\rightarrow4} \rightarrow Conv_{4\rightarrow2}$) to parameterize the stochastic latent $U$. We set $\alpha=10^{-4}$ and $\beta=0.0003$. 

For the lightweight \textbf{HiDDeN}, the VIB module is integrated via a Multi-Layer Perceptron (MLP) structure between the feature extractor and the readout layer, utilizing a latent dimension of $D=128$. The hyperparameters are set to $\alpha=0.007$ and $\beta=0.00015$. 
Detailed network architectures, layer configurations, and specific implementation parameters are provided in \textbf{Appendix \ref{app:implementation}}.

\subsection{Zero-Shot Resilience to AIGC Manipulation}

\begin{table}[ht]
\centering
\caption{\textbf{Zero-shot Robustness against Generative Editing.} 
Evaluation under (I) Local Editing and (II) Global Purification. 
The reported metric is Bit Error Rate (BER). 
\textbf{Red.} denotes the relative reduction in error rate.}
\label{tab:aigc_robustness}
\small
\setlength{\tabcolsep}{4.5pt}
\begin{tabular}{lccc}
\toprule
\textbf{Attack Method} & \textbf{EditGuard} & \textbf{+VIB (Ours)} & \textbf{Red.} \\
\midrule

\multicolumn{4}{l}{\textit{\textbf{\textcolor{gray}{(I) Local Editing (BER \textperthousand)}}}} \\
SD-Inpainting        & 0.35 & \textbf{0.03} & \textbf{\textcolor{cvprblue}{91\%}} \\
ControlNet-Inp.     & 0.13 & 0.14          & -7\%          \\
SDXL-Refiner        & 0.30 & \textbf{0.03} & \textbf{\textcolor{cvprblue}{90\%}} \\
RePaint             & 0.26 & \textbf{0.08} & \textbf{\textcolor{cvprblue}{69\%}} \\
\midrule
\multicolumn{4}{l}{\textit{\textbf{\textcolor{gray}{(II) Global Purification (BER \%)}}}} \\
SD-v1.5             & 48.55 & \textbf{38.34} & \textbf{\textcolor{cvprblue}{21\%}} \\
SD-2-Inpainting     & 48.69 & \textbf{35.21} & \textbf{\textcolor{cvprblue}{28\%}} \\
ControlNet-Inp.     & 48.85 & \textbf{38.41} & \textbf{\textcolor{cvprblue}{21\%}} \\
DDPM-CelebA-HQ      & 61.45 & \textbf{20.20} & \textbf{\textcolor{cvprblue}{67\%}} \\
SDXL                & 25.90 & \textbf{14.62} & \textbf{\textcolor{cvprblue}{44\%}} \\
SDXL-Inpainting     & 25.79 & \textbf{13.19} & \textbf{\textcolor{cvprblue}{49\%}} \\
\bottomrule
\multicolumn{4}{l}{\scriptsize \textbf{Attack strength settings:}} \\
\multicolumn{4}{l}{\scriptsize SD-v1.5 / SD-2 / ControlNet / DDPM: $0.002$; SDXL / SDXL-Inpainting: $0.05$.}\\
\multicolumn{4}{l}{\scriptsize PSNR is around 30db for SD 2.0 Purification and 16db for SD 1.5 ones}

\end{tabular}
\end{table}

To verify the defense against \textbf{Generative Purification}, we evaluate the models under two distinct zero-shot threat models: \textbf{Localized Semantic Editing} and \textbf{Global Generative Purification}. The results are summarized in \textbf{Table \ref{tab:aigc_robustness}}.

On the AGE-Set localized AIGC tampering dataset~\cite{zhang2024editguard}, integrating WaterVIB suppresses the average BER from 0.26\textperthousand to \textbf{0.07\textperthousand}, achieving a \textbf{73\% relative reduction}. Specifically, against powerful generators like SD-Inpainting~\cite{rombach2022high} and SDXL-Refiner~\cite{podell2023sdxl}, WaterVIB reduces the error rate by over 90\%. Furthermore, our method demonstrates \textbf{consistent generalization} across diverse architectures, maintaining robust extractability on RePaint~\cite{lugmayr2022repaint} and ControlNet-Inpainting~\cite{zhang2023adding}.

We further evaluate robustness under \textbf{global AIGC purification} using a diverse set of generative models.
As shown in Table~\ref{tab:aigc_robustness}, global reconstruction substantially degrades the baseline across all models, with BER ranging from 25\% to over 60\%.
EditGuard-VIB consistently reduces BER, achieving up to 67\% relative improvement under pixel-space DDPM and nearly 50\% under strong SDXL-based purification.

We note that the relatively high BER on SD-v1.5 is primarily caused by its limited reconstruction fidelity (PSNR $\approx$ 15\,dB), which induces severe pixel-level distortions and consequently strong watermark corruption.
This behavior is expected for text-conditioned latent diffusion models optimized for perceptual realism rather than pixel accuracy.

\subsection{Robustness against Standard Distortions}
\label{sec:standard_robustness}

While our primary focus is generative defense, an ideal watermarking scheme must also excel at standard signal processing benchmarks. In this section, we verify the efficacy of WaterVIB across both high-capacity and lightweight architectures.

\begin{table}[h]
\centering
\caption{\textbf{Comparison with SOTA on Standard Distortions.} Evaluation on EditGuard(+VIB) architecture. We report PSNR (dB) for imperceptibility and Bit Error Rate (BER \%) for robustness.}
\label{tab:sota_standard}
\small
\setlength{\tabcolsep}{5pt}
\begin{tabular}{lcccc}
\toprule
\textbf{Method} & \textbf{Payload} & \textbf{PSNR} $\uparrow$ & \textbf{Clean}\textperthousand $\downarrow$ & \textbf{Noise}\textperthousand $\downarrow$ \\
\midrule
DWT-DCT & 100 bits & 38.1 & 0.56 & 10.84 \\
TrustMark & 100 bits & \textbf{42.3} & 1.00 & 6.60 \\
WM-A & 30 bits & 38.6  & 0.63  & 140.20 \\
EditGuard & 100 bits & 40.4  & 0.05 & 3.21 \\
\midrule
\textbf{+VIB (Ours)} & \textbf{100 bits} & 40.3 & \textbf{0.03} & \textbf{0.08} \\
\bottomrule
\end{tabular}
\end{table}

%\paragraph{High-Capacity SOTA Comparison.}
\textbf{High-Capacity SOTA Comparison.} We benchmark EditGuard-VIB against recent SoTA methods. As shown in \textbf{Table \ref{tab:sota_standard}}, under random attacks (specifically Gaussian, Poisson, and JPEG), integrating WaterVIB reduces the BER from 3.21\textperthousand to \textbf{0.08\textperthousand}. This performance consistently outperforms current SOTA models, including TrustMark~\citep{bui2023trustmark} (6.60\% BER) and WM-A~\citep{sander2025watermark}.

\begin{table}[h]
\centering
\caption{\textbf{Detailed Robustness Analysis (EditGuard).} Comparison of Bit Error Rate (BER \%). \textbf{Red.} denotes the relative reduction in error rate.}
\label{tab:editguard_detailed}
%\small
% 稍微调小列间距以适应新增的一列
\setlength{\tabcolsep}{4pt}
\begin{tabular}{lccc}
\toprule
\textbf{Distortion Type} & \textbf{Baseline} & \textbf{+VIB(Ours)} & \textbf{Red.} \\
\midrule
Gaussian Noise\textperthousand & 0.09 & \textbf{0.01} & \textbf{\textcolor{cvprblue}{89\%}} \\
JPEG \textperthousand & 9.61 & \textbf{0.40} & \textbf{\textcolor{cvprblue}{96\%}} \\
Poisson Noise\textperthousand & 0.01 & 0.02 & - \\
Cropout\% & 29.61 & 32.82 & - \\
Dropout\% & 9.11 & \textbf{4.50} &\textbf{ \textcolor{cvprblue}{51\%}} \\
\textbf{Resize (Scaling)\%} & 81.75 & \textbf{0.01} & \textbf{\textcolor{cvprblue}{99.99\%}} \\
\midrule
\textit{Combined Average\%} & \textit{20.24} & \textit{\textbf{6.23}} & \textit{\textbf{\textcolor{cvprblue}{69\%}}} \\
\bottomrule
\end{tabular}
\end{table}

As detailed in \textbf{Table \ref{tab:editguard_detailed}}, the baseline suffers a catastrophic collapse under \textbf{Resize} (81.75\% BER), betraying its dependency on position-specific pixel grids. WaterVIB virtually eliminates this vulnerability (\textbf{0.01\%} BER), confirming that the information bottleneck enforces strict invariance to grid resampling.

\begin{table}[ht]
\centering
\caption{\textbf{Universality on Lightweight Models (HiDDeN).} Evaluation on HiDDeN (30 bits). We report PSNR (dB) and Bit Error Rate (BER \%). \textbf{Red.} denotes the relative reduction in error rate. Note the significant improvements across structural (Resize/Cropout) and adversarial attacks.}
\label{tab:hidden_universality}
\small
\begin{tabular}{lccc}
\toprule
\textbf{Metric } & \textbf{Baseline} & \textbf{+VIB(Ours)} & \textbf{Red.} \\
\midrule
\multicolumn{4}{l}{\textit{\textbf{\textcolor{gray}{(I) Visual Quality \& Clean Accuracy}}}} \\
PSNR (dB) $\uparrow$ & 36.5 & \textbf{37.0} & - \\
SSIM$\uparrow$  & 0.97 & \textbf{0.97}  &  -      \\
Clean BER (\%) & 9.92 & \textbf{9.54} & \textcolor{cvprblue}{4\%} \\
\midrule
\multicolumn{4}{l}{\textit{\textbf{\textcolor{gray}{(II) Standard Distortions}}}} \\
Crop (0.5) & 10.37 & \textbf{9.76} & \textcolor{cvprblue}{6\%} \\
Cropout (0.3) & 24.95 & \textbf{9.69} & \textbf{\textcolor{cvprblue}{61\%}} \\
Dropout (0.3) & 16.10 & \textbf{12.46} & \textcolor{cvprblue}{23\%} \\
JPEG  & 29.01 & \textbf{12.45} & \textbf{\textcolor{cvprblue}{57\%}} \\
Resize (Scaling) & 32.04 & \textbf{12.93} & \textbf{\textcolor{cvprblue}{60\%}} \\
\textit{Combined Average} & \textit{20.92} & \textit{\textbf{13.02}} & \textit{\textcolor{cvprblue}{38\%}} \\
\midrule
\multicolumn{4}{l}{\textit{\textbf{\textcolor{gray}{(III) Complex \& Adversarial Attacks}}}} \\
Geometric$^{\dagger}$ & 24.00 & \textbf{15.00} & \textbf{\textcolor{cvprblue}{38}}\% \\
Color Space (YUV) & 16.00 & \textbf{11.00} & \textcolor{cvprblue}{31\%} \\
Frequency (DWT) & 44.00 & \textbf{39.00} & \textcolor{cvprblue}{11\%} \\
Adversarial (PGD) & 76.00 & \textbf{30.00} & \textbf{\textcolor{cvprblue}{61\%}} \\
\bottomrule
\multicolumn{4}{l}{\scriptsize $^{\dagger}$ Rot+Scale+Trans+Shear.} \\
\vspace*{-0.5cm}
\end{tabular}
\end{table}

%\paragraph{Robustness on Lightweight Models.}
\textbf{Robustness on Lightweight Models.} To demonstrate universality, we evaluate WaterVIB on the lightweight HiDDeN architecture. As detailed in \textbf{Table \ref{tab:hidden_universality}}, WaterVIB exhibits particular resilience against \textbf{distribution-shifting distortions}, such as complex geometric attacks (\textbf{Red. 38\%}) and JPEG compression (\textbf{Red. 57\%}). This strong generalization capability extends even to aggressive PGD adversarial perturbations, where WaterVIB prevents the baseline's collapse (76.0\% $\to$ 30.0\% BER). Collectively, these findings validate that the VIB module acts as a potent regularizer for parameter-constrained networks, forcing them to capture robust semantics rather than overfitting to fragile details.

\subsection{Analysis of the Information Sieve Mechanism}
\label{sec:mechanism_analysis}

To uncover the underlying mechanisms of WaterVIB's robustness, we visualize the feature space shifts under AIGC tampering and further quantify the gradient interference. This analysis confirms that WaterVIB effectively resolves the theoretical vulnerabilities identified in Section \ref{sec:theory}.

\begin{figure}[ht]
\centering
\includegraphics[width=0.98\linewidth]{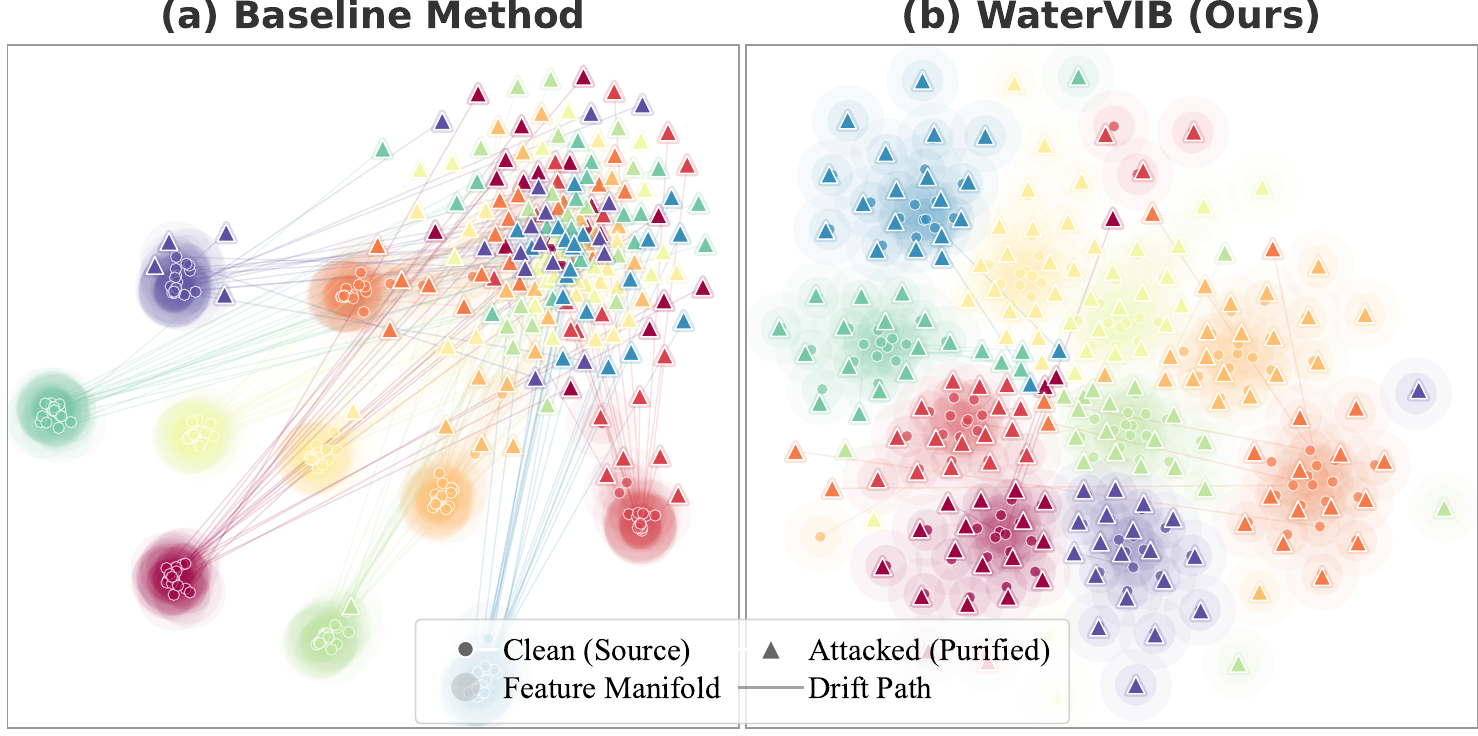}
\caption{\textbf{Feature Space Visualization (t-SNE).} We visualize the latent embeddings of 10 random messages (different colors), each embedded into 20 cover images. In the Baseline (EditGuard), attacked samples (triangles) undergo significant feature drift, collapsing toward a shared manifold region regardless of their original message identity, which leads to high bit error rates.
(b) With our WaterVIB, the drift paths (lines) are significantly reduced, and attacked samples remain anchored within the high-density clusters of their respective clean counterparts. }
\label{fig:tsne_vis}
\end{figure}

\textbf{Feature Space Invariance (t-SNE).}
We first verify whether the Information Sieve successfully disentangles the watermark from fragile cover distortions. We randomly select 10 watermark messages and embed each into 20 distinct cover images. We then visualize the latent representations extracted by the decoder for these samples using t-SNE. In \textbf{Figure \ref{fig:tsne_vis}}, we project both the clean samples and their corresponding AIGC-purified counterparts into the same 2D embedding space.

As shown in Figure \ref{fig:tsne_vis} (Top), the baseline exhibits a severe \textit{distribution shift}. While clean samples form distinct clusters, the purified samples (triangles) drift significantly away from their class centers and converge into a new cluster, indicating that the encoder relies on texture-dependent features that are effectively rewritten by the generative editing process.

In contrast, WaterVIB (Bottom) demonstrates remarkable \textit{manifold invariance}. The attacked samples remain tightly clustered with their corresponding clean anchors. This qualitative result confirms that the VIB module filters out the ``nuisance'' variability, forcing the model to learn a MSS representation that is structurally invariant to regeneration.

% --- Insert Figure 3 Here ---

\textbf{Mitigating Gradient Interference.}
We further provide a quantitative explanation based on the ``Gradient Counter-Optimization'' theory proposed in Section \ref{subsec:collapse}. We measure the \textbf{Gradient Interference Ratio ($\rho$)}, defined as the ratio between the projection of the attack distortion $s_{atk}$ and the watermark signal $s_{wm}$ onto the decoding gradient $\nabla_x \mathcal{L}$:
\begin{equation}
    \rho = \frac{\langle s_{atk}, \nabla_x \mathcal{L} \rangle}{|\langle s_{wm}, \nabla_x \mathcal{L} \rangle|}
\end{equation}
A higher $\rho$ implies that generative purification acts as a stronger adversarial attack that cancels out more watermark signal.

%In stark contrast, integrating WaterVIB suppresses this ratio to \textbf{0.1167}, achieving a significant \textbf{73\% reduction}. This quantitative drop validates that the Information Sieve mechanism forces the encoder to learn a representation that is statistically distinct from the manifold projections performed by diffusion models. Effectively, WaterVIB pushes the watermark signal into a subspace that is nearly \textit{orthogonal} to the purification trajectory, thereby \textbf{substantially mitigating the ``Gradient Counter-Optimization'' effect.}

% --- Insert Table 7 Here ---
\begin{table}[h]
\centering
\caption{\textbf{Gradient Interference Analysis.} Comparison of the Gradient Interference Ratio ($\rho$). Lower is better. 
% WaterVIB significantly reduces the adversarial interference by 73\%.
}
\label{tab:gradient_interference}
\small
\begin{tabular}{lccc}
\toprule
\textbf{Method} & \textbf{EditGuard} & \textbf{+VIB (Ours)} & \textbf{Red.} \\
\midrule
Ratio \textbf{$\rho$} $\downarrow$ & 0.4285 & \textbf{0.1167} & \textbf{73\%} \\
\bottomrule
\end{tabular}
\end{table}

As reported in \textbf{Table \ref{tab:gradient_interference}}, the baseline suffers from high interference ($\rho \approx 0.43$), confirming that purification acts as a direct adversarial update. In contrast, WaterVIB suppresses this ratio by \textbf{73\%} (to 0.1167). This validates that the Information Sieve forces the watermark into a latent subspace that achieves \textit{quasi-orthogonality} relative to the purification trajectory, thereby effectively mitigating the ``Gradient Counter-Optimization'' effect at a structural level.

More details are presented in Appendix \ref{app:supp_experiments}

% As reported in \textbf{Table \ref{tab:gradient_interference}}, the baseline suffers from a high interference ratio of \textbf{0.50} (1/2), meaning the attack effectively neutralizes half of the decoding signal. WaterVIB significantly reduces this ratio to \textbf{0.33} (1/3). This suggests that by constraining the information flow, WaterVIB forces the watermark into a subspace that is more \textit{orthogonal} to the manifold projection direction, thereby reducing the adversarial impact of the purification process.

\subsection{Training Dynamics and Generalization}
\label{sec:dynamics}

\begin{figure}[ht]
\centering
\vspace*{-0.2cm}
% \hspace*{-0.6cm}
\includegraphics[width=\linewidth]{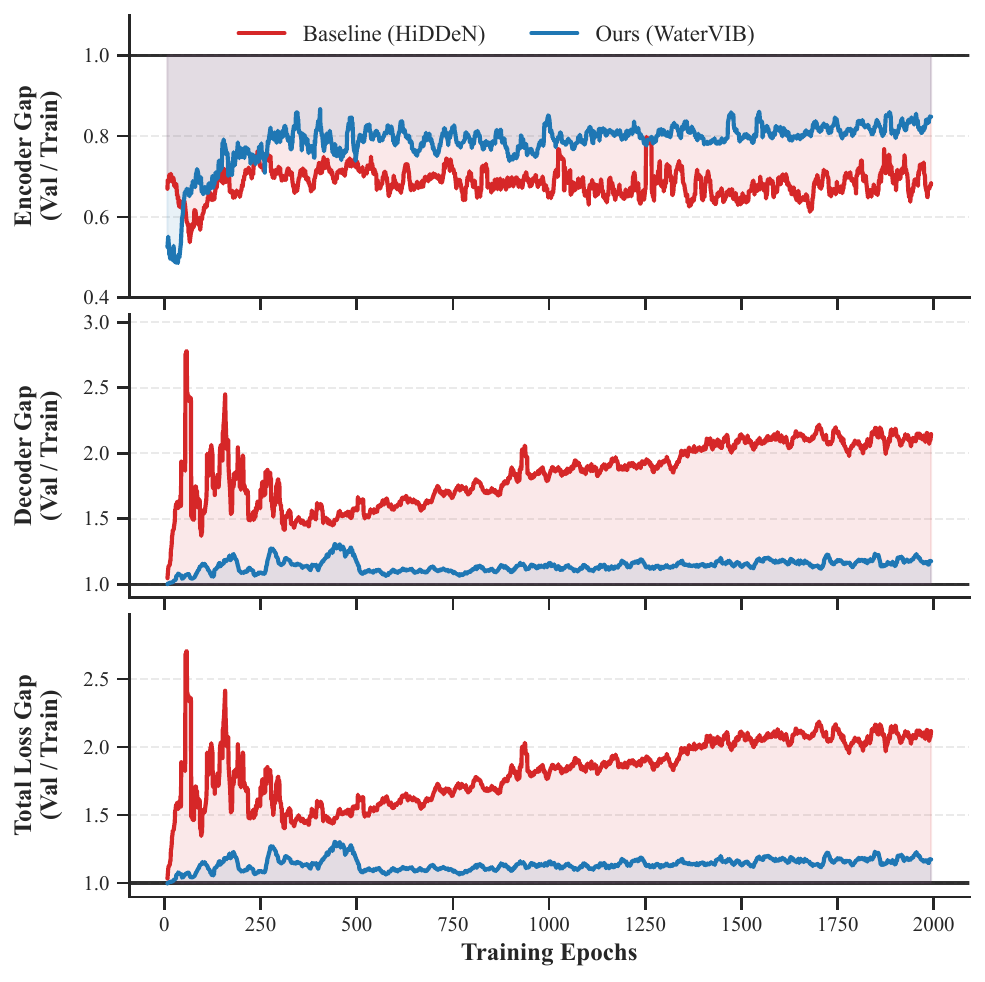}
%\caption{\textbf{Training Dynamics.} Comparison of Train vs. Val BER curves. \textbf{Left:} Baseline (HiDDeN). \textbf{Right:} WaterVIB (Ours).}
\vspace{-12pt}
\caption{\textbf{Generalization Gap Analysis.} We evaluate the training dynamics by plotting the ratio of validation loss to training loss ($\mathcal{L}_{\text{val}} / \mathcal{L}_{\text{train}}$) across epochs. A ratio significantly greater than $1$ indicates overfitting. The plots correspond to the Encoder (top), Decoder (middle), and Total Loss (bottom). 
%While the Baseline (red) exhibits severe overfitting in the decoding stage, our WaterVIB (blue) consistently maintains a ratio near $1.0$, demonstrating superior generalization and training stability.
}

\label{fig:training_dynamics}
\end{figure}

To further verify the regularization effect of WaterVIB, we analyze the learning curves and the behavior of the regularization term $L_{\textbf{KL}}$ during training. \textbf{Figure \ref{fig:training_dynamics}} visualizes the Train vs. Val Loss for both the Baseline and WaterVIB.

%\textbf{Reduced Generalization Gap.} The Baseline (Figure \ref{fig:training_dynamics} Left) exhibits a large generalization gap where training error vanishes while validation error remains high, indicating overfitting to training-specific artifacts. Conversely, WaterVIB (Right) significantly narrows this gap with the validation curve closely tracking training performance. This confirms that the Information Bottleneck acts as an effective regularizer, filtering out non-robust nuisance correlations to ensure generalization.

%\textbf{Stable Compression Intensity.} Furthermore, we analyze the evolution of the KL divergence loss ($L_{KL}$). Empirically, we observe that after the first 20 epochs, $L_{KL}$ stabilizes at approximately \textbf{0.0005 ($\pm$ 0.0001)}. This numerical stability confirms that the VIB module exerts a constant regularization pressure, consistently and stably filtering out task-irrelevant features to enforce feature minimality throughout the training.

\textbf{Reduced Generalization Gap.} We observe that the baseline’s embedding intensity and decoding accuracy on the validation set is consistently lower than its training performance (Figure \ref{fig:training_dynamics} red). We attribute this to the model's embedding strategy overfitting to the specific textures and local patterns of the training set. Consequently, when deployed on unseen test data, these novel textures confuse the model, leading to a significant drop in encoder intensity and a corresponding increase in decoder gap loss. In contrast, WaterVIB significantly avoids dependency on specific textures, making the model's performance more stable. This phenomenon validates our design objective, confirming that the watermarking strategy reduces its reliance on original cover textures and leads to superior generalization performance.

\textbf{Stable Compression.} The KL divergence ($L_{KL}$) stabilizes at $\approx \textbf{0.0005}$ after 20 epochs. This numerical stability indicates that the VIB module exerts constant regularization pressure, consistently filtering out task-irrelevant features.

\subsection{Hyperparameter Ablation}
\label{sec:sensitivity}

As formally proven in \textbf{Appendix B}, the hyperparameter $\beta$ determines the compression rate of the Information Sieve. It effectively sets the ``aperture'' of the bottleneck: a higher $\beta$ enforces a stricter constraint, compelling the encoder to discard more task-irrelevant information. To empirically determine the optimal operating point, we conducted a systematic ablation study on the HiDDeN backbone.

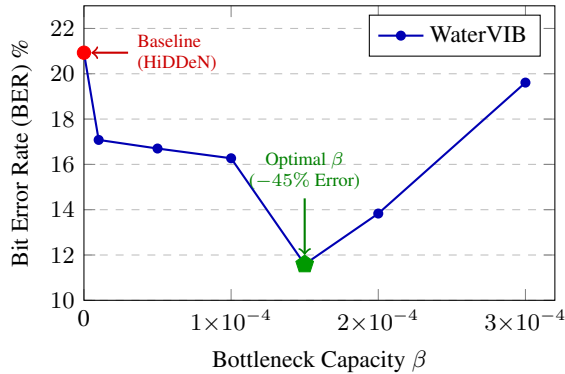
\begin{figure}[th]
\centering
\begin{tikzpicture}
\begin{axis}[
    width=0.95\linewidth,
    height=5.5cm,
    xlabel={Bottleneck Capacity $\beta$},
    ylabel={Bit Error Rate (BER) \%},
    xmin=0, xmax=0.00032,
    ymin=10, ymax=23,
    xtick={0, 0.0001, 0.0002, 0.0003},
    xticklabels={0, $1{\times}10^{-4}$, $2{\times}10^{-4}$, $3{\times}10^{-4}$},
    ytick={10, 12, 14, 16, 18, 20, 22},
    ymajorgrids=true,
    grid style=dashed,
    legend pos=north east,
    legend style={font=\small},
    label style={font=\small},
    tick label style={font=\small},
    scaled x ticks=false
]

% 1. 绘制主曲线
\addplot[
    color=blue!70!black,
    mark=*,
    thick,
    mark size=1.5pt
]
coordinates {
    (0, 20.93)
    (0.00001, 17.08)
    (0.00005, 16.70)
    (0.0001, 16.27)
    (0.00015, 11.59)
    (0.0002, 13.83)
    (0.0003, 19.61)
};
\addlegendentry{WaterVIB}

% 2. 标注 Baseline (红点)
\addplot[only marks, mark=*, mark size=2.5pt, color=red] coordinates {(0, 20.93)};

% === 修改处：Baseline 文字移到正右方 ===
% 坐标设为 (0.00003, 20.93)，即高度不变，水平拉开
% anchor=west 意味着以文字框的“左边”为定位点，这样文字自然向右延伸
\node[anchor=west, text=red!80!black, font=\scriptsize, align=left] (base_text) at (axis cs:0.00003, 20.93) {Baseline\\(HiDDeN)};
% 箭头从文字的西边(west) 指向 点的边缘 (0.000005)
\draw[->, red!80!black, thick] (base_text.west) -- (axis cs:0.000005, 20.93);

% 3. 标注 Optimal Point (绿色五边形)
\addplot[only marks, mark=pentagon*, mark size=3.5pt, color=green!60!black] coordinates {(0.00015, 11.59)};

% Optimal 文字 (保持在上方)
\node[anchor=south, text=green!50!black, font=\scriptsize, align=center] (opt_text) at (axis cs:0.00015, 14.5) {Optimal $\beta$\\($-45\%$ Error)};
\draw[->, green!50!black, thick] (opt_text.south) -- (axis cs:0.00015, 12.0);

\end{axis}
\end{tikzpicture}
\vspace*{-0.2cm}
\caption{\textbf{Impact of $\beta$ on Robustness.} We plot the BER under AIGC purification as a function of $\beta$. The Baseline coresponds to $\beta=0$. The curve reveals a distinct ``sweet spot'' at $\beta=0.00015$.}
\label{fig:beta_ablation}
\end{figure}

\textbf{Figure \ref{fig:beta_ablation}} visualizes the results, where we treat the standard HiDDeN model as the unconstrained special case ($\beta=0$). As shown in the plot, the performance exhibits a distinct \textbf{non-monotonic ``U-shaped'' pattern}.

 \textbf{Transition from Baseline ($\beta=0 \to 1.5{\times}10^{-4}$):} Initially, increasing $\beta$ significantly drops the BER from the baseline's 20.93\% to a minimum of \textbf{11.59\%}. This \textbf{45\% reduction} confirms that introducing the VIB constraint successfully filters out redundant, texture-dependent features that are vulnerable to generative attacks.
 
 \textbf{Over-compression ($\beta > 2.0{\times}10^{-4}$):} However, beyond the optimal point, the bottleneck becomes excessively tight. The penalty on mutual information $I(Z;X)$ begins to discard essential watermark signals, causing the BER to rise sharply.

This analysis identifies $\beta \approx 1.5{\times}10^{-4}$ as the ``sweet spot'' that optimally balances feature minimality with information sufficiency, with the method consistently outperforming the baseline across the entire stable range.

\section{Conclusion}
In this work, we identify feature entanglement as the critical failure mode of deep watermarking under generative purification. To resolve this, we bridge the gap between robust watermarking and Information-Theoretic Representation Learning. We prove that optimizing the Information Bottleneck objective is equivalent to learning a Minimal Sufficient Statistic (MSS) of the message within our framework, which constitutes a necessary condition for disentangling the watermark from fragile cover textures. By operationalizing this theory via WaterVIB, we achieve SoTA zero-shot robustness against a wide spectrum of unknown AIGC purifications. 
Crucially, Our results substantiate that the VIB framework constitutes a potent defense paradigm against zero-shot attacks. 
Our work suggests that future defenses should move beyond heuristic noise layers toward theoretically grounded, semantic-invariant representation learning.

%%%%%%%%%%%%%%%%%%%%%%%%%%%%%%%%%%%%%%%%%%%%%%%%%%%%%%%%%%%%%%%%%%%%%%%%%%%%%%%%%%%%%%%%%%%%%%%%%%%%%%%%%%%%%%%%%%%%%%%%%%%%%%%%%%%%%%%%%%%%%%%%%%%%%%%%%%%%%%%%%%%%%%%%%%%%%%

\section*{Impact Statement}

This paper presents WaterVIB, a robust watermarking framework designed to safeguard intellectual property and ensure content provenance in the era of generative AI. Our work aims to empower creators by providing a defense against unauthorized manipulation and erasure of copyright information. While watermarking technologies inherently carry potential risks related to user tracking or surveillance if misused, our research focuses strictly on the robustness of signal retention for ownership verification. We are committed to the ethical development of these tools and advocate for their use solely in protecting legitimate copyright and ensuring digital authenticity.


\bibliography{main}

\begin{thebibliography}{30}
\providecommand{\natexlab}[1]{#1}
\providecommand{\url}[1]{\texttt{#1}}
\expandafter\ifx\csname urlstyle\endcsname\relax
  \providecommand{\doi}[1]{doi: #1}\else
  \providecommand{\doi}{doi: \begingroup \urlstyle{rm}\Url}\fi

\bibitem[Alemi et~al.(2016)Alemi, Fischer, Dillon, and Murphy]{alemi2016deep}
Alemi, A.~A., Fischer, I., Dillon, J.~V., and Murphy, K.
\newblock Deep variational information bottleneck.
\newblock \emph{arXiv preprint arXiv:1612.00410}, 2016.

\bibitem[Boyd \& Vandenberghe(2004)Boyd and Vandenberghe]{Boyd2004ConvexO}
Boyd, S. and Vandenberghe, L.
\newblock \emph{Convex optimization}.
\newblock Cambridge university press, 2004.

\bibitem[Bui et~al.(2023)Bui, Agarwal, and Collomosse]{bui2023trustmark}
Bui, T., Agarwal, S., and Collomosse, J.
\newblock Trustmark: Universal watermarking for arbitrary resolution images.
\newblock \emph{arXiv preprint arXiv:2311.18297}, 2023.

\bibitem[Ganic \& Eskicioglu(2004)Ganic and Eskicioglu]{ganic2004robust}
Ganic, E. and Eskicioglu, A.~M.
\newblock Robust dwt-svd domain image watermarking: embedding data in all frequencies.
\newblock In \emph{MM and Sec Workshop}, pp.\  166--174, 2004.

\bibitem[Gilad-Bachrach et~al.(2003)Gilad-Bachrach, Navot, and Tishby]{GiladBachrach2003AnIT}
Gilad-Bachrach, R., Navot, A., and Tishby, N.
\newblock An information theoretic tradeoff between complexity and accuracy.
\newblock In \emph{Learning Theory and Kernel Machines: 16th Annual Conference on Learning Theory and 7th Kernel Workshop, COLT/Kernel 2003, Washington, DC, USA, August 24-27, 2003. Proceedings}, pp.\  595--609. Springer, 2003.

\bibitem[Jing et~al.(2021)Jing, Deng, Xu, Wang, and Guan]{jing2021hinet}
Jing, J., Deng, X., Xu, M., Wang, J., and Guan, Z.
\newblock Hinet: Deep image hiding by invertible network.
\newblock In \emph{CVPR}, pp.\  4733--4742, 2021.

\bibitem[Kang et~al.(2003)Kang, Huang, Shi, and Lin]{kang2003dwt}
Kang, X., Huang, J., Shi, Y.~Q., and Lin, Y.
\newblock A dwt-dft composite watermarking scheme robust to both affine transform and jpeg compression.
\newblock \emph{TCSVT}, 13\penalty0 (8):\penalty0 776--786, 2003.

\bibitem[Kingma \& Welling(2013)Kingma and Welling]{kingma2013auto}
Kingma, D.~P. and Welling, M.
\newblock Auto-encoding variational bayes.
\newblock \emph{arXiv preprint arXiv:1312.6114}, 2013.

\bibitem[Langelaar \& Lagendijk(2001)Langelaar and Lagendijk]{langelaar2001optimal}
Langelaar, G.~C. and Lagendijk, R.~L.
\newblock Optimal differential energy watermarking of dct encoded images and video.
\newblock \emph{TIP}, 10\penalty0 (1):\penalty0 148--158, 2001.

\bibitem[Lehmann \& Scheff{\'e}(2011)Lehmann and Scheff{\'e}]{lehmann2011completeness}
Lehmann, E.~L. and Scheff{\'e}, H.
\newblock Completeness, similar regions, and unbiased estimation-part i.
\newblock In \emph{Selected works of EL Lehmann}, pp.\  233--268. Springer, 2011.

\bibitem[Lin et~al.(2014)Lin, Maire, Belongie, Hays, Perona, Ramanan, Doll{\'a}r, and Zitnick]{lin2014microsoft}
Lin, T.-Y., Maire, M., Belongie, S., Hays, J., Perona, P., Ramanan, D., Doll{\'a}r, P., and Zitnick, C.~L.
\newblock Microsoft coco: Common objects in context.
\newblock In \emph{ECCV}, pp.\  740--755. Springer, 2014.

\bibitem[Lugmayr et~al.(2022)Lugmayr, Danelljan, Romero, Yu, Timofte, and Van~Gool]{lugmayr2022repaint}
Lugmayr, A., Danelljan, M., Romero, A., Yu, F., Timofte, R., and Van~Gool, L.
\newblock Repaint: Inpainting using denoising diffusion probabilistic models.
\newblock In \emph{Proceedings of the IEEE/CVF conference on computer vision and pattern recognition}, pp.\  11461--11471, 2022.

\bibitem[Luo et~al.(2025)Luo, Rocha, Shi, Guo, Li, and Wan]{luo2025nerf}
Luo, Z., Rocha, A., Shi, B., Guo, Q., Li, H., and Wan, R.
\newblock The nerf signature: Codebook-aided watermarking for neural radiance fields.
\newblock \emph{TPAMI}, 2025.

\bibitem[Mildenhall et~al.(2019)Mildenhall, Srinivasan, Ortiz-Cayon, Kalantari, Ramamoorthi, Ng, and Kar]{mildenhall2019local}
Mildenhall, B., Srinivasan, P.~P., Ortiz-Cayon, R., Kalantari, N.~K., Ramamoorthi, R., Ng, R., and Kar, A.
\newblock Local light field fusion: Practical view synthesis with prescriptive sampling guidelines.
\newblock \emph{ACM Transactions on Graphics (ToG)}, 38\penalty0 (4):\penalty0 1--14, 2019.

\bibitem[Mildenhall et~al.(2021)Mildenhall, Srinivasan, Tancik, Barron, Ramamoorthi, and Ng]{mildenhall2021nerf}
Mildenhall, B., Srinivasan, P.~P., Tancik, M., Barron, J.~T., Ramamoorthi, R., and Ng, R.
\newblock Nerf: Representing scenes as neural radiance fields for view synthesis.
\newblock \emph{Communications of the ACM}, 65\penalty0 (1):\penalty0 99--106, 2021.

\bibitem[Nie et~al.(2022)Nie, Guo, Huang, Xiao, Vahdat, and Anandkumar]{nie2022diffusion}
Nie, W., Guo, B., Huang, Y., Xiao, C., Vahdat, A., and Anandkumar, A.
\newblock Diffusion models for adversarial purification.
\newblock \emph{arXiv preprint arXiv:2205.07460}, 2022.

\bibitem[Podell et~al.(2023)Podell, English, Lacey, Blattmann, Dockhorn, M{\"u}ller, Penna, and Rombach]{podell2023sdxl}
Podell, D., English, Z., Lacey, K., Blattmann, A., Dockhorn, T., M{\"u}ller, J., Penna, J., and Rombach, R.
\newblock Sdxl: Improving latent diffusion models for high-resolution image synthesis.
\newblock \emph{arXiv preprint arXiv:2307.01952}, 2023.

\bibitem[Razeghi et~al.(2022)Razeghi, Rezaeifar, Ferdowsi, Holotyak, and Voloshynovskiy]{razeghi2022compressed}
Razeghi, B., Rezaeifar, S., Ferdowsi, S., Holotyak, T., and Voloshynovskiy, S.
\newblock Compressed data sharing based on information bottleneck model.
\newblock In \emph{ICASSP}, pp.\  3009--3013. IEEE, 2022.

\bibitem[Rombach et~al.(2022)Rombach, Blattmann, Lorenz, Esser, and Ommer]{rombach2022high}
Rombach, R., Blattmann, A., Lorenz, D., Esser, P., and Ommer, B.
\newblock High-resolution image synthesis with latent diffusion models.
\newblock In \emph{CVPR}, pp.\  10684--10695, 2022.

\bibitem[Sander et~al.(2025)Sander, Fernandez, Durmus, Furon, and Douze]{sander2025watermark}
Sander, T., Fernandez, P., Durmus, A., Furon, T., and Douze, M.
\newblock Watermark anything with localized messages.
\newblock In \emph{ICLR}, 2025.

\bibitem[Seo et~al.(2023)Seo, Kim, and Park]{seo2023interpretable}
Seo, S., Kim, S., and Park, C.
\newblock Interpretable prototype-based graph information bottleneck.
\newblock \emph{NeurIPS}, 36:\penalty0 76737--76748, 2023.

\bibitem[Tancik et~al.(2020)Tancik, Mildenhall, and Ng]{tancik2020stegastamp}
Tancik, M., Mildenhall, B., and Ng, R.
\newblock Stegastamp: Invisible hyperlinks in physical photographs.
\newblock In \emph{CVPR}, pp.\  2117--2126, 2020.

\bibitem[Tishby et~al.(2000)Tishby, Pereira, and Bialek]{tishby2000information}
Tishby, N., Pereira, F.~C., and Bialek, W.
\newblock The information bottleneck method.
\newblock \emph{arXiv preprint physics/0004057}, 2000.

\bibitem[Yang et~al.(2025)Yang, Qi, Ren, Jia, Sun, Zhu, and Liao]{yang2025exploring}
Yang, Z., Qi, Z., Ren, Z., Jia, Z., Sun, H., Zhu, X., and Liao, X.
\newblock Exploring information processing in large language models: Insights from information bottleneck theory.
\newblock \emph{arXiv preprint arXiv:2501.00999}, 2025.

\bibitem[Zhang et~al.(2020)Zhang, Benz, Karjauv, Sun, and Kweon]{zhang2020udh}
Zhang, C., Benz, P., Karjauv, A., Sun, G., and Kweon, I.~S.
\newblock Udh: Universal deep hiding for steganography, watermarking, and light field messaging.
\newblock \emph{Advances in Neural Information Processing Systems}, 33:\penalty0 10223--10234, 2020.

\bibitem[Zhang et~al.(2023)Zhang, Rao, and Agrawala]{zhang2023adding}
Zhang, L., Rao, A., and Agrawala, M.
\newblock Adding conditional control to text-to-image diffusion models.
\newblock In \emph{Proceedings of the IEEE/CVF international conference on computer vision}, pp.\  3836--3847, 2023.

\bibitem[Zhang et~al.(2024)Zhang, Li, Yu, Xu, Li, and Zhang]{zhang2024editguard}
Zhang, X., Li, R., Yu, J., Xu, Y., Li, W., and Zhang, J.
\newblock Editguard: Versatile image watermarking for tamper localization and copyright protection.
\newblock In \emph{CVPR}, pp.\  11964--11974, 2024.

\bibitem[Zhao et~al.(2024)Zhao, Zhang, Su, Vasan, Grishchenko, Kruegel, Vigna, Wang, and Li]{Zhao2023InvisibleIW}
Zhao, X., Zhang, K., Su, Z., Vasan, S., Grishchenko, I., Kruegel, C., Vigna, G., Wang, Y.-X., and Li, L.
\newblock Invisible image watermarks are provably removable using generative ai.
\newblock \emph{Advances in neural information processing systems}, 37:\penalty0 8643--8672, 2024.

\bibitem[Zhong et~al.(2023)Zhong, Wang, Yao, Hu, Dong, and Munteanu]{zhong2023semantic}
Zhong, R., Wang, R., Yao, W., Hu, M., Dong, S., and Munteanu, A.
\newblock Semantic representation and attention alignment for graph information bottleneck in video summarization.
\newblock \emph{TIP}, 32:\penalty0 4170--4184, 2023.

\bibitem[Zhu et~al.(2018)Zhu, Kaplan, Johnson, and Fei-Fei]{zhu2018hidden}
Zhu, J., Kaplan, R., Johnson, J., and Fei-Fei, L.
\newblock Hidden: Hiding data with deep networks.
\newblock In \emph{ECCV}, pp.\  657--672, 2018.

\end{thebibliography}
\bibliographystyle{icml2026}

%%%%%%%%%%%%%%%%%%%%%%%%%%%%%%%%%%%%%%%%%%%%%%%%%%%%%%%%%%%%%%%%%%%%%%%%%%%%%%%
%%%%%%%%%%%%%%%%%%%%%%%%%%%%%%%%%%%%%%%%%%%%%%%%%%%%%%%%%%%%%%%%%%%%%%%%%%%%%%%
% APPENDIX
%%%%%%%%%%%%%%%%%%%%%%%%%%%%%%%%%%%%%%%%%%%%%%%%%%%%%%%%%%%%%%%%%%%%%%%%%%%%%%%
%%%%%%%%%%%%%%%%%%%%%%%%%%%%%%%%%%%%%%%%%%%%%%%%%%%%%%%%%%%%%%%%%%%%%%%%%%%%%%%
\newpage
\appendix
\onecolumn
\section{Theoretical Equivalence to Classical Minimal Sufficient Statistics}
\label{sec:appendix_proof}

In this section, we provide a rigorous justification for our information-theoretic definition of the Minimal Sufficient Statistic (MSS). Under the assumption that the representation mapping $T(\cdot)$ is a \textbf{deterministic function} (consistent with neural network inference) and the data is defined on \textbf{discrete domains}, we demonstrate a bidirectional equivalence between the classical definition of MSS grounded in the Lehmann-Scheff\'{e} theory~\cite{lehmann2011completeness} and the proposed information-theoretic formulation.

\subsection{Preliminaries}
We consider the standard supervised setting for watermark extraction defined over discrete alphabets:
\begin{itemize}
    \item \textbf{Message (Target):} $M \sim p(m)$, representing the discrete watermark message.
    \item \textbf{Observation (Data):} $X \sim p(x|m)$, representing the watermarked image (discrete random variable, e.g., digitized pixels).
    \item \textbf{Statistic (Representation):} Let $Z$ be a statistic of $X$, defined as a deterministic function $Z = T(X)$.
\end{itemize}

We assume the data generation and feature extraction process follows a Markov Chain:
\begin{equation}
    M \to X \to Z
\end{equation}

\subsection{Classical Statistical Definitions}
We first restate the foundational definitions from classical mathematical statistics adapted for discrete variables.

\vspace{0.5em}
\noindent\textbf{Definition 1 (Sufficiency).}
A statistic $T(X)$ is sufficient for $M$ if the conditional distribution of $X$ given $T(X)$ is independent of $M$. That is, for all $x, t, m$:
\begin{equation}
    p(x | t, m) = p(x | t)
\end{equation}

\vspace{0.5em}
\noindent\textbf{Definition 2 (Minimal Sufficiency via Lehmann-Scheff\'{e}).}
A sufficient statistic $T(X)$ is called a Minimal Sufficient Statistic (MSS) if, for any other sufficient statistic $S(X)$, $T(X)$ is a function of $S(X)$. That is, there exists a function $f$ such that:
\begin{equation}
    T(X) = f(S(X)) \quad \text{almost surely.}
\end{equation}
This definition implies that the MSS induces the coarsest sufficient partition of the sample space.

\subsection{Information-Theoretic Foundations}
To prove the main theorems, we first establish precise properties of deterministic functions and sufficiency using entropy and mutual information definitions for discrete variables.

\vspace{0.5em}
\noindent\textbf{Proposition 1 (Vanishing Conditional Entropy for Deterministic Functions).}
\textit{For any discrete random variable $X$ and a deterministic function $T(\cdot)$, the conditional entropy $H(T(X)|X)$ is zero.}

\noindent\textit{Proof.} Let $Z = T(X)$. By the definition of conditional entropy for discrete variables:
\begin{equation}
    H(Z|X) = - \sum_{x \in \mathcal{X}} p(x) \sum_{z \in \mathcal{Z}} p(z|x) \log p(z|x)
\end{equation}
Since $T$ is a deterministic function, the conditional probability mass function $p(z|x)$ is an indicator function: it equals 1 if $z = T(x)$ and 0 otherwise. The entropy of a deterministic event is zero ($-1 \log 1 = 0$). Thus, the inner sum vanishes for all $x$:
\begin{equation}
    \sum_{z \in \mathcal{Z}} p(z|x) \log p(z|x) = 0
\end{equation}
Averaging over $p(x)$ preserves this zero value, yielding $H(Z|X) = 0$. \hfill $\square$

\vspace{0.5em}
\noindent\textbf{Proposition 2 (Equivalence of Conditional Independence and Zero Information).}
\textit{The random variables $X$ and $M$ are conditionally independent given $T(X)$ if and only if the conditional mutual information $I(M; X | T(X))$ is zero.}

\noindent\textit{Proof.}
The conditional mutual information is defined as the expected Kullback-Leibler (KL) divergence:
\begin{equation}
    I(M; X | T) = \mathbb{E}_{t} \left[ D_{KL}(p(m, x | t) \| p(m|t)p(x|t)) \right]
\end{equation}
\textbf{Direction ($\Rightarrow$):} If $X$ and $M$ are conditionally independent given $T$, then $p(m, x | t) = p(m|t)p(x|t)$. The KL divergence between two identical distributions is zero, thus the term vanishes.

\noindent\textbf{Direction ($\Leftarrow$):} A fundamental property of KL divergence (Gibbs' Inequality) states that $D_{KL}(P\|Q) \ge 0$, with equality if and only if $P = Q$. Therefore, if $I(M; X | T) = 0$, it implies $p(m, x | t) = p(m|t)p(x|t)$ for all $(m,x,t)$ with non-zero probability. This factorization is the definition of conditional independence. \hfill $\square$

\vspace{0.5em}
\noindent\textbf{Theorem 1 (Information-Theoretic Criterion for Sufficiency).}
\textit{A statistic $T(X)$ is sufficient for $M$ if and only if it preserves the mutual information between the input and the target:}
\begin{equation}
    I(M; T(X)) = I(M; X)
\end{equation}

\noindent\textit{Proof.}
We analyze the joint mutual information $I(M; X, T(X))$ using the Chain Rule. Since $T(X)$ is deterministic given $X$, $H(T(X)|X)=0$ (Proposition 1), which implies $I(M; T(X) | X) = 0$. Thus:
\begin{equation}
    I(M; X, T(X)) = I(M; X)
\end{equation}
Alternatively, expanding in the reverse order:
\begin{equation}
    I(M; X, T(X)) = I(M; T(X)) + I(M; X | T(X))
\end{equation}
Combining these, we get the fundamental identity:
\begin{equation}
    I(M; X) = I(M; T(X)) + I(M; X | T(X))
\end{equation}
From this identity, the equality $I(M; X) = I(M; T(X))$ holds if and only if $I(M; X | T(X)) = 0$. By \textbf{Proposition 2}, $I(M; X | T(X)) = 0$ is equivalent to conditional independence ($M \perp X | T(X)$), which is precisely the definition of Sufficiency. \hfill $\square$

\vspace{0.5em}
\noindent\textbf{Lemma 1 (Information Invariance under Deterministic Mapping).}
\textit{Let $Z$ be a discrete random variable and $Y = g(Z)$ be a deterministic function of $Z$. The mutual information with any source $X$ is preserved in the joint pair $(Z, Y)$: $I(X; Z) = I(X; Z, g(Z))$.}

\noindent\textit{Proof.} Applying the Chain Rule: $I(X; Z, Y) = I(X; Z) + I(X; Y | Z)$. Since $Y$ is deterministic given $Z$, $H(Y|Z)=0$, implying $I(X; Y | Z) = 0$. Thus, $I(X; Z, g(Z)) = I(X; Z)$. \hfill $\square$

\vspace{0.5em}
\noindent\textbf{Lemma 2 (Invertibility from Information Equality).}
\textit{Let $Z$ be a random variable determined by $X$. Let $T^* = g(Z)$. If $I(X; Z) = I(X; T^*)$, then $g$ is invertible on the support of $Z$.}

\noindent\textit{Proof.} Define information loss $\Delta I = I(X; Z) - I(X; T^*)$. By \textbf{Lemma 1}, $I(X; Z) = I(X; Z, T^*) = I(X; T^*) + I(X; Z | T^*)$. Thus, $\Delta I = I(X; Z | T^*)$.
Given $I(X; Z) = I(X; T^*)$, we have $I(X; Z | T^*) = 0$. Written as expected KL divergence:
\begin{equation}
    \mathbb{E}_{t^*} \left[ D_{KL}(p(x, z | t^*) \| p(x|t^*)p(z|t^*)) \right] = 0
\end{equation}
By Gibbs' Inequality, this implies $p(x, z | t^*) = p(x|t^*)p(z|t^*)$.
Since $Z$ is a deterministic function of $X$ (say $Z = T_{enc}(X)$), the conditional PMF is $p(z|x, t^*) = \mathbb{I}(z = T_{enc}(x))$, where $\mathbb{I}$ is the indicator function. The conditional independence implies $p(z|x, t^*) = p(z|t^*)$.
Therefore, $p(z|t^*) = \mathbb{I}(z = T_{enc}(x))$, meaning $Z$ is fully determined by $T^*$. This proves the existence of an inverse mapping $Z = f(T^*)$, making $g$ bijective on the support. \hfill $\square$

\subsection{Theorem 2: Classical MSS Implies Information Minimality}

\noindent\textbf{Theorem 2.}
\textit{Let $T(X)$ be a Minimal Sufficient Statistic for $M$ in the classical sense. Let $S(X)$ be any other sufficient statistic for $M$. Then, $T(X)$ minimizes the mutual information with the input $X$ among all sufficient statistics:}
\begin{equation}
    I(T(X); X) \le I(S(X); X)
\end{equation}

\noindent\textit{Proof.}
Let $T(X)$ be the Minimal Sufficient Statistic and $S(X)$ be an arbitrary sufficient statistic for $M$. By the definition of minimal sufficiency (Definition 2), $T(X)$ must be a function of $S(X)$, denoted as $T(X) = f(S(X))$. This functional relationship establishes a Markov Chain $X \to S(X) \to T(X)$.

Applying the Data Processing Inequality (DPI) to this chain, we obtain $I(X; T(X)) \le I(X; S(X))$. Since $T(X)$ is sufficient, it retains all mutual information regarding the target $M$ (\textbf{Theorem 1}), while the DPI confirms it simultaneously minimizes the information retained about the input $X$ compared to any other sufficient statistic $S(X)$. \hfill $\blacksquare$

\subsection{Theorem 3: Information Minimality Implies Classical MSS}

\noindent\textbf{Theorem 3 (Converse).}
\textit{Let $Z$ be a sufficient statistic for $M$ such that for any other sufficient statistic $S(X)$, $I(Z; X) \le I(S; X)$. Let $T^*(X)$ be a classical Minimal Sufficient Statistic. Then, $Z$ is isomorphic to $T^*(X)$.}

\noindent\textit{Proof.}
Assume a classical MSS $T^*(X)$ exists. Since $Z$ is hypothesized to be a sufficient statistic, by the definition of classical minimal sufficiency, $T^*(X)$ must be a function of $Z$. Let this mapping be $T^*(X) = g(Z)$. This dependency forms the Markov Chain $X \to Z \to T^*(X)$, and by the Data Processing Inequality, we have $I(X; T^*) \le I(X; Z)$.

Conversely, the theorem hypothesis states that $Z$ minimizes mutual information among \textit{all} sufficient statistics. Since $T^*(X)$ is itself a sufficient statistic, it must hold that $I(X; Z) \le I(X; T^*)$. Combining these two inequalities, we obtain the equality:
\begin{equation}
    I(X; Z) = I(X; T^*) = I(X; g(Z))
\end{equation}
Invoking \textbf{Lemma 2}, the information equality $I(X; Z) = I(X; g(Z))$ implies that the function $g$ is invertible on the support of $Z$. Consequently, there exists a deterministic inverse $Z = g^{-1}(T^*)$. Since $T^*$ is a function of $Z$ and $Z$ is a function of $T^*$, $Z$ is strictly equivalent to the classical MSS $T^*$ up to isomorphism. \hfill $\blacksquare$
% You can have as much text here as you want. The main body must be at most $8$
% pages long. For the final version, one more page can be added. If you want, you
% can use an appendix like this one.

%The $\mathtt{\backslash onecolumn}$ command above can be kept in place if you
%prefer a one-column appendix, or can be removed if you prefer a two-column
%appendix.  Apart from this possible change, the style (font size, spacing,
%margins, page numbering, etc.) should be kept the same as the main body.
%%%%%%%%%%%%%%%%%%%%%%%%%%%%%%%%%%%%%%%%%%%%%%%%%%%%%%%%%%%%%%%%%%%%%%%%%%%%%%%
%%%%%%%%%%%%%%%%%%%%%%%%%%%%%%%%%%%%%%%%%%%%%%%%%%%%%%%%%%%%%%%%%%%%%%%%%%%%%%%

\section{Derivation of the Lagrangian and the $\beta$-$\epsilon$ Correspondence}
\label{sec:appendix_1_proof}

This appendix provides the formal justification for replacing the constrained $\epsilon$-Minimal Sufficient Statistic (MSS) problem with the unconstrained Lagrangian relaxation used in WaterVIB. Specifically, we prove the strict convexity of the Information Bottleneck (IB) curve for the watermarking channel and establish the bijective mapping between the Lagrange multiplier $\beta$ and the sufficiency tolerance $\epsilon$.

\subsection{The Primal Optimization Problem}

Let $X$ denote the cover signal, $M$ the watermark message, and $Z$ the stochastic representation (watermarked latent). We assume the Markov chain $M \rightarrow X \rightarrow Z$. The optimization goal is to find the encoder $p(z|x)$ that minimizes the compression rate $R$ while maintaining a relevance $I$ at least $I_{total} - \epsilon$. The primal problem $(P_\epsilon)$\cite{tishby2000information} is defined over the convex set of valid conditional distributions $\Delta$:$$\begin{aligned}
    \min_{p(z|x) \in \Delta} \quad & I(Z; X) \\
    \text{s.t.} \quad & I(Z; M) \ge I_{total} - \epsilon
\end{aligned}$$where $I_{total} = I(X; M)$.

\subsection{Lagrangian Construction}

The mutual information functionals $R(p)$ and $I(p)$ are convex and concave functions of the mapping $p(z|x)$, respectively. Since the optimization is performed over a convex set $\Delta$, we can employ the method of Lagrange multipliers to convert the constrained problem into an unconstrained variational problem.

We introduce a Lagrange multiplier $\lambda \ge 0$ associated with the inequality constraint. The Lagrangian function $\mathcal{L}(p, \lambda)$ is defined as:
\begin{equation}
    \mathcal{L}(p, \lambda) = R(p) - \lambda \left( I(p) - (I_{total} - \epsilon) \right)
\end{equation}

The optimization problem seeks to minimize this Lagrangian with respect to the encoder $p$:
\begin{equation}
    p^*_\lambda = \arg \min_{p \in \Delta} \left\{ R(p) - \lambda I(p) + \lambda(I_{total} - \epsilon) \right\}
\end{equation}

Since the term $\lambda(I_{total} - \epsilon)$ is constant with respect to $p$, the optimization simplifies to:
\begin{equation}
    \min_{p} \big( R(p) - \lambda I(p) \big) \iff \max_{p} \big( \lambda I(p) - R(p) \big)
\end{equation}

To align this formulation with the standard Information Bottleneck (IB) objective, we define the trade-off parameter $\beta \triangleq \frac{1}{\lambda}$. Assuming the constraint is active (implying $\lambda > 0$), multiplying the objective by $\frac{1}{\lambda}$ does not alter the optimal solution $p^*$. Thus, the problem is equivalent to maximizing:
\begin{equation}
\label{eq:watervib_objective_derived}
    \mathcal{L}_{\textbf{IB}}(p, \beta) = I(Z; M) - \beta I(Z; X)
\end{equation}
This confirms that maximizing the IB
objective~\citep{tishby2000information} is mathematically equivalent to solving the primal MSS problem for a specific Lagrange multiplier $\lambda = 1/\beta$.

\subsection{Step 3: Geometry of the Solution and Strict Convexity}

The relationship between the hyperparameter $\beta$ and the tolerance $\epsilon$ is governed by the geometry of the \textit{Information Plane}~\citep{GiladBachrach2003AnIT}. We define the \textit{Minimal Rate Curve} function $R_{min}(i)$, which characterizes the Pareto frontier of the compression-relevance trade-off. Specifically, $R_{min}(i)$ represents the minimum compression rate required to achieve a target relevance level $i$:
\begin{equation}
    R_{min}(i) = \min_{p \in \Delta} \{ I(Z; X) \mid I(Z; M) = i \}
\end{equation}

\textbf{Property 1 (Strict Convexity and Monotonicity).} According to the standard Information Bottleneck theory (Tishby et al., 2000), the Minimal Rate Curve function $R_{min}(i)$ is strictly convex and monotonically increasing for $i \in [0, I_{total}]$. Mathematically:
\begin{equation}
    \frac{d R_{min}(i)}{d i} > 0 \quad \text{and} \quad \frac{d^2 R_{min}(i)}{d i^2} > 0
\end{equation}

\textit{Constraint Activation.} In our primal problem, we seek to minimize the compression rate $R(p)$. Let $i_{target} = I_{total} - \epsilon$ be the lower bound of the constraint. Suppose, for the sake of contradiction, that the optimal solution $p^*$ satisfies the strict inequality $I(p^*) > i_{target}$.
Due to the monotonicity property ($\frac{d R_{min}}{d i} > 0$), a reduction in relevant information $I(p)$ leads to a strict reduction in the minimal required rate $R_{min}$. Therefore, there exists another encoder $p'$ with $I(p') = i_{target}$ such that $R(p') < R(p^*)$. This contradicts the optimality of $p^*$.
Consequently, the optimal solution must lie exactly on the boundary of the feasible region:
\begin{equation}
    I(p^*) = I_{total} - \epsilon
\end{equation}

\subsection{Step 4: Proof of the Bijective Mapping ($\beta \leftrightarrow \epsilon$)}

We now establish the one-to-one correspondence between $\beta$ and $\epsilon$ using the Karush-Kuhn-Tucker (KKT) conditions. In convex optimization~\citep{Boyd2004ConvexO}, the optimal Lagrange multiplier $\lambda$ represents the sensitivity of the objective function's optimal value to perturbations in the constraint bound (often referred to as the \textit{shadow price}).

Specifically, let $p^*(i)$ denote the optimal encoder for a given relevance target $i = I_{total} - \epsilon$. According to the \textbf{Stationarity} condition of KKT, the gradient of the Lagrangian with respect to the distribution $p$ must vanish at the optimum:
\begin{equation}
    \nabla_p \mathcal{L}(p^*, \lambda) = \nabla_p R(p^*) - \lambda \nabla_p I(p^*) = 0 \implies \nabla_p R(p^*) = \lambda \nabla_p I(p^*)
\end{equation}

Now, consider the Minimal Rate Curve $R_{min}(i) = R(p^*(i))$. We analyze how the minimal rate changes with an infinitesimal perturbation in the target relevance $i$. By applying the chain rule:
\begin{equation}
    \frac{d R_{min}(i)}{d i} = \left\langle \nabla_p R(p^*), \frac{d p^*}{d i} \right\rangle
\end{equation}
Substituting the stationarity condition $\nabla_p R(p^*) = \lambda \nabla_p I(p^*)$ into the equation:
\begin{equation}
    \frac{d R_{min}(i)}{d i} = \left\langle \lambda \nabla_p I(p^*), \frac{d p^*}{d i} \right\rangle = \lambda \underbrace{\left\langle \nabla_p I(p^*), \frac{d p^*}{d i} \right\rangle}_{\frac{d I(p^*)}{d i}}
\end{equation}
Since the constraint is active (as proven in Step 3), we have $I(p^*(i)) = i$. Differentiating both sides with respect to $i$ yields $\frac{d I(p^*)}{d i} = 1$.
Consequently, the relationship simplifies directly to:
\begin{equation}
    \frac{d R_{min}(i)}{d i} = \lambda \cdot 1 = \lambda
\end{equation}
This confirms the geometric interpretation of the Lagrange multiplier $\lambda$ as the slope of the Minimal Rate Curve.

Formally, at the optimal solution corresponding to a tolerance $\epsilon$, we have:
\begin{equation}
    \lambda(\epsilon) = \frac{\partial R_{min}(i)}{\partial i} \bigg|_{i = I_{total} - \epsilon}
\end{equation}

Substituting our definition $\beta = 1/\lambda$, we derive the explicit mapping function:
\begin{equation}
\label{eq:beta_epsilon_map}
    \beta(\epsilon) = \left( \frac{\partial R_{min}(i)}{\partial i} \bigg|_{i = I_{total} - \epsilon} \right)^{-1} = \frac{d i}{d R_{min}} \bigg|_{R_{min} = R^*}
\end{equation}
This equation states that $\beta$ is the inverse of the slope of the Minimal Rate Curve at the point determined by $\epsilon$.

\textbf{Theorem (Bijective Correspondence).} The mapping $\mathcal{T}: \epsilon \to \beta$ defined in Eq.~\ref{eq:beta_epsilon_map} is a bijection for $\epsilon \in (0, I_{total})$.

\textit{Proof.}
\begin{enumerate}
    \item \textbf{Monotonicity:} Since $R_{min}(i)$ is strictly convex (Property 1), its first derivative $\lambda(i) = \frac{d R_{min}}{d i}$ is strictly monotonically increasing with respect to the relevance $i$. Consequently, as $\epsilon$ increases (meaning relevance $i = I_{total} - \epsilon$ decreases), the slope $\lambda$ strictly decreases.
    \item \textbf{Invertibility:} A strictly monotonic function is inherently invertible. Therefore, for every specific tolerance $\epsilon$, there exists a unique slope $\lambda$ defining the tangent to the curve at that point. Since $\beta = 1/\lambda$, there is a unique $\beta$ corresponding to each $\epsilon$.
\end{enumerate}

\textbf{Conclusion.} This proof demonstrates that adjusting the hyperparameter $\beta$ in the WaterVIB objective (Eq.~\ref{eq:watervib_objective_derived}) is mathematically equivalent to traversing the Pareto frontier of the $\epsilon$-MSS problem. A larger $\beta$ implies a steeper slope on the rate-relevance curve, corresponding to a looser tolerance $\epsilon$ (more compression, less information retained), whereas a smaller $\beta$ enforces a stricter $\epsilon$ (higher fidelity to the watermark).

\newpage

\section{Appendix C: Implementation Details}
\label{app:implementation}

In this section, we provide the parameter-level details of the WaterVIB framework, including the specific network architectures for the Information Sieve mechanism, training hyperparameters, and the configuration of the differentiable noise layers used during the attack simulation phase to ensure reproducibility.

\subsection{The WaterVIB Module Design}

The WaterVIB module functions as a stochastic information bottleneck inserted between the feature extractor and the message decoder. To ensure numerical stability and gradient flow during end-to-end optimization, we implement the module using the reparameterization trick. Specifically, the latent variable $U$ is computed as
\begin{equation}
\label{eq:detail_U_design}
    U = \mu(Z) + \alpha \cdot \epsilon \odot \exp(\frac{1}{2}\log\sigma(Z)^2)
\end{equation}, where $\epsilon \sim \mathcal{N}(0, I)$ is sampled from a standard normal distribution. Here, $\alpha$ is a scalar scaling factor introduced to control the variance magnitude during the early stages of training. To prevent numerical instability such as exploding gradients, we explicitly clip the predicted log-variance $\log\Sigma(Z)^2$ to the range $[-10, 10]$ before the exponential operation.

The training objective is a composite loss function:
\begin{equation}
    \mathcal{L}_{total} = \lambda_{img}\mathcal{L}_{img} + \lambda_{rec}\mathcal{L}_{rec} + \beta\mathcal{L}_{KL}
\end{equation} The reconstruction loss $\mathcal{L}_{rec}$ is calculated as the Binary Cross-Entropy (BCE) between the ground truth message and the predicted probabilities, while the compression loss $\mathcal{L}_{KL}$ is the Kullback-Leibler divergence between the posterior $q(U|Z)$ and the prior $r(U) = \mathcal{N}(0, I)$, weighted by the bottleneck capacity $\beta$. The image fidelity loss $\mathcal{L}_{img}$ measures the Mean Squared Error (MSE) between the cover and watermarked images.

\subsection{Integration into High-Capacity Architecture (EditGuard)}

For the EditGuard backbone, which utilizes a high-capacity encoder-decoder structure, we insert the VIB module after the 16-channel output of the bit decoder to handle the spatial feature maps. The input to the VIB module is a feature map of shape $[B, 16, 400, 400]$. To handle this high dimensionality efficiently, we employ a CNN-based compression pipeline rather than fully connected layers. The features are first downsampled and compressed via a sequence of channel reduction layers: a $Conv2d(16 \to 4)$ layer reduces the channel dimension while downsampling the spatial resolution to $128 \times 128$, followed by a $Conv2d(4 \to 2)$ layer. A final convolution layer $Conv2d(2 \to 2)$ produces the distributional parameters $\mu$ and $\log\Sigma^2$, each maintaining the shape $[B, 1, 128, 128]$. After sampling the stochastic latent $U$, the feature map is flattened to a dimension of $16,384$ and projected via a single Linear layer ($16,384 \to 100$) to produce the 100-bit output logits.

Regarding hyperparameters, we trained the model on the COCO2017 dataset at a resolution of $512 \times 512$. We set the bottleneck capacity $\beta$ to $0.0003$ and the noise scaling factor $\alpha$ to $10^{-4}$ to mitigate fluctuations during the initial training phase.

\subsection{Integration into Lightweight Architecture (HiDDeN)}

For the parameter-constrained HiDDeN backbone, the VIB module is integrated between the final convolutional block of the extractor and the linear readout layer. Unlike the CNN-based approach used in EditGuard, this implementation utilizes a Multi-Layer Perceptron (MLP) structure. The global features from the extractor are mapped to a latent dimension of $D=128$. Two parallel linear layers then map these features to the statistical parameters $\mu \in \mathbb{R}^{128}$ and $\log\Sigma^2 \in \mathbb{R}^{128}$. The sampled latent $U$ is finally mapped to the 30-bit output via a linear readout layer.

This model was trained on 10,000 images from the COCO2014 dataset at a resolution of $256 \times 256$. Based on our ablation studies, we determined the optimal bottleneck capacity $\beta$ to be $0.00015$ and set the noise scaling factor $\alpha$ to $0.007$.

\subsection{Differentiable Noise Simulation and Training}

To ensure robust generalization, we train the models against a comprehensive suite of differentiable noise layers. For EditGuard-VIB, the training focused on Gaussian, Poisson, and JPEG noise to strictly evaluate zero-shot generalization capabilities against unseen AIGC attacks. As shown in Table \ref{tab:noise_params}, the parameters define the range of random distortions applied on HiDDeN-VIB during the attack simulation.
% For the HiDDeN-VIB experiments, the noise simulation includes random cropping with a ratio $p \in [0.4, 0.55]$, Cropout masking with $p \in [0.25, 0.35]$, Dropout with a probability $p \in [0.65, 0.75]$, and Resizing with a scaling factor $s \in [0.4, 0.6]$. We also employ a differentiable approximation of JPEG compression with coefficients $(25, 9, 9)$. 

\begin{table}[h]
\centering
\caption{Configuration of Differentiable Noise Layers used in HiDDeN-VIB Training.}
\label{tab:noise_params}
\begin{tabular}{lcc}
\toprule
\textbf{Distortion Type} & \textbf{Parameter} & \textbf{Range / Value} \\
\midrule
Crop & Ratio & $[0.4, 0.55]$ \\
Cropout & Ratio & $[0.25, 0.35]$ \\
Dropout & Keep Prob. & $[0.65, 0.75]$ \\
Resize (Scaling) & Scale Factor & $[0.4, 0.6]$ \\
JPEG Compression & Coefficients & $(25, 9, 9)$ \\
\bottomrule
\end{tabular}
\end{table}

All models were implemented in PyTorch and trained on NVIDIA 4090 GPUs using the Adam optimizer. The learning rate was initialized at $10^{-3}$ and adjusted using a scheduler. We used a batch size of 32 for the lightweight HiDDeN architecture and 64 for the high-capacity EditGuard architecture.

\subsection{Training Algorithm}
\label{app:training_algo}

We summarize the end-to-end training process of the WaterVIB framework in Algorithm \ref{algo:watervib_training}. This algorithm details the stochastic information bottleneck mechanism and the optimization of the composite objective function.

\begin{algorithm}[h]
   \caption{Training Process of WaterVIB}
   \label{algo:watervib_training}
\begin{algorithmic}[1]
   \STATE {\bfseries Input:} Training dataset $\mathcal{D}$, Batch size $B$, Learning rate $\eta$, Hyperparameters $\alpha, \beta$
   \STATE {\bfseries Initialize:} Encoder $E$, Extractor $E_{ext}$, MLPs ($\text{MLP}_\mu, \text{MLP}_\sigma$), Decoder $D_{msg}$ parameters $\theta$
   \FOR{each training epoch}
   \FOR{each batch $(x, m)$ in $\mathcal{D}$}
   \STATE \textit{\# 1. Watermark Embedding}
   \STATE $x_{wm} \leftarrow E(x, m)$ \COMMENT{Generate watermarked image}
   
   \STATE \textit{\# 2. Attack Simulation (Differentiable)}
   \STATE $x_{atk} \leftarrow \text{Attack}(x_{wm})$ \COMMENT{Apply noise layers (e.g., Crop, JPEG)}
   
   \STATE \textit{\# 3. Deterministic Feature Extraction}
   \STATE $Z \leftarrow E_{ext}(x_{atk})$ \COMMENT{Extract backbone features}
   
   \STATE \textit{\# 4. Stochastic Bottleneck (VIB)}
   \STATE $\mu \leftarrow \text{MLP}_\mu(Z)$
   \STATE $\sigma \leftarrow \exp\left(\frac{1}{2}\text{MLP}_\sigma(Z)\right)$
   \STATE Sample $\epsilon \sim \mathcal{N}(0, I)$
   \STATE $U \leftarrow \mu + \alpha \cdot \epsilon \odot \sigma$ \COMMENT{Reparameterization Trick (Eq. \ref{eq:detail_U_design})}
   
   \STATE \textit{\# 5. Message Decoding}
   \STATE $\hat{m} \leftarrow D_{msg}(U)$ \COMMENT{Predict message from stochastic latent}
   
   \STATE \textit{\# 6. Loss Computation}
   \STATE $\mathcal{L}_{rec} \leftarrow \text{BCE}(m, \hat{m})$ \COMMENT{Reconstruction Loss (Eq. \ref{eq:L_rec})}
   \STATE $\mathcal{L}_{KL} \leftarrow D_{KL}(\mathcal{N}(\mu, \sigma^2) \| \mathcal{N}(0, I))$ \COMMENT{Compression Loss (Eq. \ref{eq:vib_bound})}
   \STATE $\mathcal{L}_{img} \leftarrow \text{MSE}(x, x_{wm})$ \COMMENT{Image Fidelity Loss}
   \STATE $\mathcal{L}_{total} \leftarrow \mathcal{L}_{img} + \mathcal{L}_{rec} + \beta \mathcal{L}_{KL}$ \COMMENT{Total Objective (Eq. 9)}
   
   \STATE \textit{\# 7. Optimization}
   \STATE Update $\theta \leftarrow \theta - \eta \nabla_\theta \mathcal{L}_{total}$
   \ENDFOR
   \ENDFOR
\end{algorithmic}
\end{algorithm}

\newpage
\section{Appendix D: Experiments on NeRF-Signature}
\label{app:nerf_experiments}

To demonstrate the universality of our method beyond 2D static images, we integrated the WaterVIB module into NeRF-Signature~\cite{luo2025nerf}, a state-of-the-art watermarking framework for 3D Neural Radiance Fields (NeRF). This appendix details the experimental setup and the performance gains achieved by our method in the 3D domain.

\subsection{Experimental Setup}

\textbf{Datasets and Protocol.} Following the baseline NeRF-Signature framework, we evaluated our method on two standard datasets: the synthetic \textbf{Blender} dataset~\cite{mildenhall2021nerf} (Lego, Hotdog, etc.) at its native resolution, and the real-world \textbf{LLFF} dataset~\cite{mildenhall2019local} (Fern, Room, etc.) downsampled to $1/8$ resolution.

\textbf{Implementation Details.} We embed a 32-bit watermark message into the NeRF representation. The baseline NeRF-Signature injects a compact signature codebook into the radiance field. We integrated our WaterVIB module into the extraction head of the NeRF-Signature pipeline to optimize the information flow. To control randomness, all experiments employed identical NeRF cores, and evaluations were conducted on a fixed set of test viewpoints.

\textbf{Metrics.} We evaluate \textbf{imperceptibility} by comparing \textit{Watermarked} vs. \textit{Clean NeRF} renderings using PSNR, SSIM, and LPIPS~\cite{luo2025nerf}. \textbf{Robustness} is measured via Bit Accuracy (ACC) under standard 2D distortions, including Rotation, Scaling, Blurring, Brightness, JPEG, and Cropping.

\subsection{Performance Evaluation}

\textbf{Quantitative Results (Imperceptibility).}
As shown in Table~\ref{tab:nerf_clean}, the integration of WaterVIB significantly improves the visual quality of the watermarked renderings compared to the baseline. On the Blender dataset, WaterVIB increases the PSNR from 57.17 dB to 58.99 dB and reduces the LPIPS perceptual distance by 40\%, indicating that our method introduces significantly fewer artifacts while maintaining a 0\% Bit Error Rate (BER). A similar trend is observed on the real-world LLFF dataset.

\begin{table}[h]
\centering
\caption{Comparison of watermarking imperceptibility on NeRF-Signature. We measure the quality of watermarked renderings against clean NeRF renderings. Metrics include Bit Error Rate (BER, lower is better), PSNR (higher is better), LPIPS (lower is better), and SSIM (higher is better).}
\label{tab:nerf_clean}
\begin{small}
%\begin{sc}
\begin{tabular}{l|cccc|cccc}
\toprule
\multirow{2}{*}{Method} & \multicolumn{4}{c|}{\textbf{Blender Dataset}} & \multicolumn{4}{c}{\textbf{LLFF Dataset}} \\
 & BER$\downarrow$ & PSNR$\uparrow$ & LPIPS$\downarrow$ & SSIM$\uparrow$ & BER$\downarrow$ & PSNR$\uparrow$ & LPIPS$\downarrow$ & SSIM$\uparrow$ \\
\midrule
NeRF-Sig & 0.00 & 57.17 & $5 \times 10^{-5}$ & 0.9998 & 0.10 & 46.51 & 0.0016 & 0.9973 \\
\textbf{+ WaterVIB} & \textbf{0.00} & \textbf{58.99} & \textbf{$3 \times 10^{-5}$} & \textbf{0.9999} & \textbf{0.10} & \textbf{49.61} & \textbf{0.0009} & \textbf{0.9984} \\
\bottomrule
\end{tabular}
%\end{sc}
\end{small}
\end{table}

\textbf{Quantitative Results (Robustness).}
We further evaluated the robustness of the 3D watermark against 2D image distortions. As detailed in Table~\ref{tab:nerf_robustness}, WaterVIB maintains the robustness of the baseline (ACC $\approx$ 100\%) across all noise categories. W

aterVIB significantly improves imperceptibility, consistently delivering higher PSNR scores under all tested distortions (e.g., +1.3 dB in Cropping). This result confirms that the Information Sieve mechanism effectively concentrates the watermark signal into robust yet imperceptible subspaces, thereby enhancing visual fidelity without compromising extraction accuracy.

\begin{table}[h]
\centering
\caption{Robustness evaluation on the Blender dataset (32 bits). We report the Bit Accuracy (ACC, $\uparrow$) and the PSNR ($\uparrow$) of the watermarked views. WaterVIB maintains high robustness while consistently improving visual quality.}
\label{tab:nerf_robustness}
\resizebox{\textwidth}{!}{
\begin{small}
%\begin{sc}
\begin{tabular}{l|cc|cc|cc|cc|cc|cc|cc}
\toprule
\multirow{2}{*}{Method} & \multicolumn{2}{c|}{Rotation} & \multicolumn{2}{c|}{Scaling} & \multicolumn{2}{c|}{Blurring} & \multicolumn{2}{c|}{Brightness} & \multicolumn{2}{c|}{JPEG} & \multicolumn{2}{c|}{Cropping} & \multicolumn{2}{c}{Average} \\
 & Acc & PSNR & Acc & PSNR & Acc & PSNR & Acc & PSNR & Acc & PSNR & Acc & PSNR & Acc & PSNR \\
\midrule
NeRF-Sig & \textbf{0.999} & 52.36 & 1.000 & 54.94 & 1.000 & 55.55 & 1.000 & 55.20 & \textbf{0.999} & 51.76 & 1.000 & 56.23 & 0.999 & 54.32 \\
\textbf{+ WaterVIB} & 0.997 & \textbf{53.26} & 1.000 & \textbf{55.76} & \textbf{1.000} & \textbf{57.07} & 1.000 & \textbf{55.98} & 0.998 & \textbf{51.93} & 1.000 & \textbf{57.54} & 0.999 & \textbf{55.26} \\
\bottomrule
\end{tabular}
%\end{sc}
\end{small}
}
\end{table}

\subsection{Qualitative Analysis}

\begin{figure}[htbp]  % h:此处, t:顶部, b:底部, p:单独页面
    \centering  % 居中
    \includegraphics[width=0.9\textwidth]{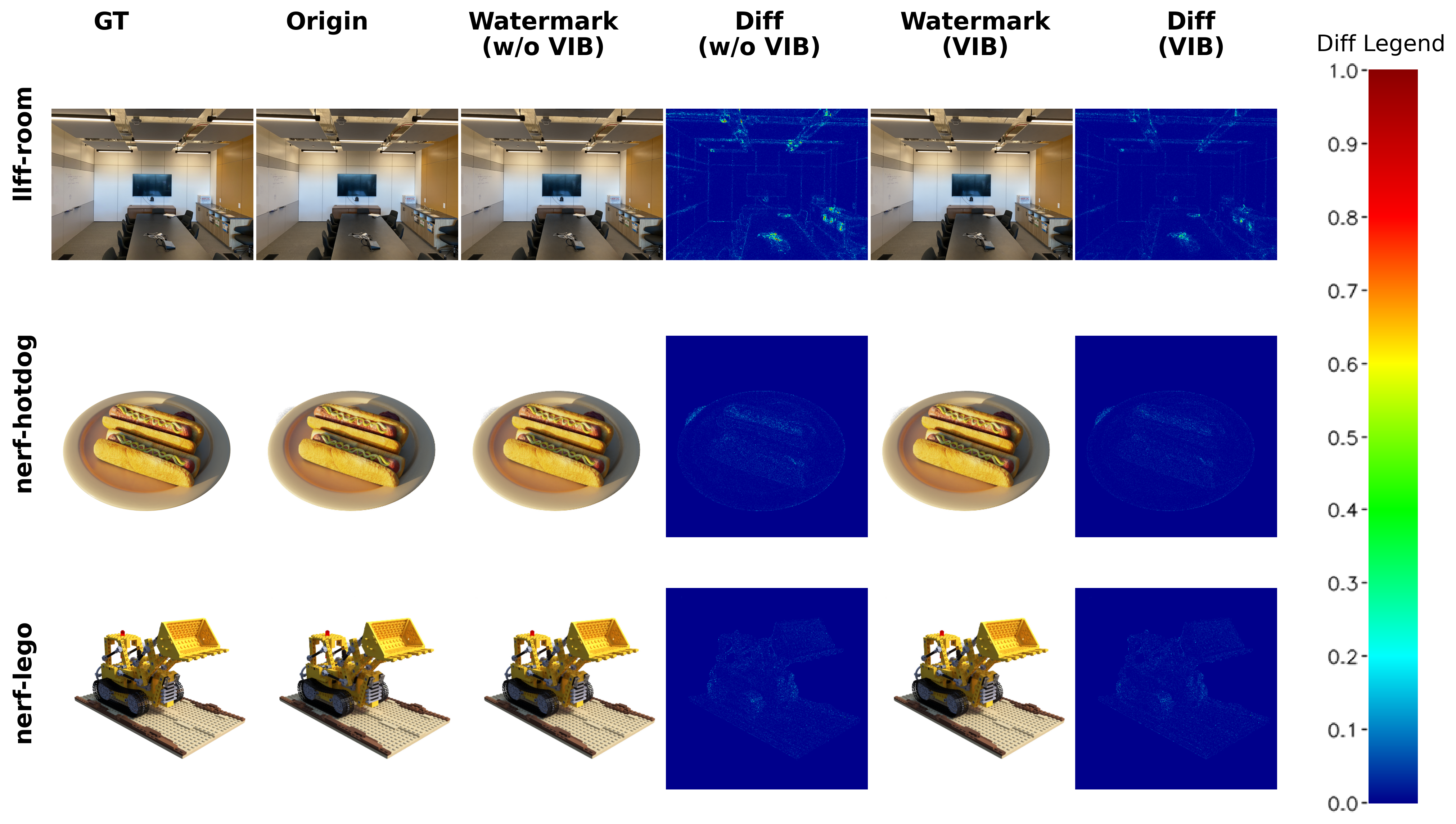}
    \caption{\textbf{Visualization of Nerf-VIB}}
    \label{fig:nerf_vis}
\end{figure}

Visual comparisons confirm that the WaterVIB module helps in concentrating the watermark signal into texture-rich, high-frequency regions of the scene, where human vision is less sensitive to perturbations. As observed in our qualitative results (Figure \ref{fig:nerf_vis}), the residual difference maps for WaterVIB are sparser and more localized compared to the baseline, which explains the simultaneous improvement in both imperceptibility (higher PSNR) and robustness.

\section{Analysis Experimental Setup and Supplement}
\label{app:supp_experiments}

In this section, we explicitly detail the experimental settings used for the theoretical validation provided in Section 3 and present additional supplementary experiments to further substantiate our findings.

\subsection{Setup for Theoretical Analysis (Section 3)}
\label{app:sec3_setup}

The theoretical analysis presented in Section \ref{sec:theory}, including the \textit{Spectral and Spatial Alignment} (Observation \ref{subsec:theory-1}) and the \textit{Gradient Counter-Optimization} analysis (Section \ref{subsec:collapse}), was conducted empirically to validate our propositions.

\textbf{Model Architecture.} All quantitative measurements and feature visualizations in Section 3 were performed using the \textbf{EditGuard} (Zhang et al., 2024) architecture as the backbone. This high-capacity model was selected to ensure that the observations regarding texture entanglement are representative of state-of-the-art deep watermarking methods.

\textbf{Data Sampling.} To strictly evaluate the statistical behavior of the watermark under attack, we randomly sampled \textbf{500 images} from the COCO validation dataset (consistent with the validation set used for EditGuard). This sample size was chosen to provide a statistically significant estimate of the spectral energy distribution and gradient projections while maintaining computational feasibility for the extensive gradient tracking required in Section 3.2.

\textbf{AIGC Purification Setting.} For the generative purification noise simulated in these analyses (specifically for the results in Tables 1, 2, and 7), we utilized the \textbf{SDXL 1.0} model. We employed the model in an image-to-image mode with a strength factor consistent with the "Global Purification" settings described in Section 5.2. This ensures that the theoretical insights regarding the "manifold projection" effect of diffusion models are directly applicable to the strongest attack scenarios evaluated in the main paper.

\textbf{Training Consistency.} The model checkpoints (CKPT) and other training hyperparameters used for these analysis steps are identical to those reported in the experimental section (Section 5 and Appendix C.2).

\subsection{Additional Supplementary Analysis Experiments}
\label{app:more_experiments}
To provide a more comprehensive analysis of WaterVIB, we conduct the following additional experiments.

\textbf{Experiment I: Geometric Orthogonality and Energy Analysis.} 
To strictly quantify the disentanglement, we analyze the geometric relationship between the watermark signal $\mathbf{s}_{wm}$ and the AIGC distortion vector $\mathbf{s}_{atk}$ on 500 randomly sampled images.

\begin{table}[h]
    \centering
    \caption{\textbf{Orthogonality Analysis.} We compare the cosine similarity and the effective projection of the attack noise onto the watermark direction. \textbf{Effective Proj.} serves as the decisive metric for signal survival.}
    \label{tab:orthogonality_analysis}
    \begin{small}
    \begin{tabular}{l|cc}
        \toprule
        \textbf{Method} & \textbf{Cosine Sim.} ($\cos \theta$) & \textbf{Effective Proj.} ($\eta \cdot |\cos \theta|$) \\
        \midrule
        Baseline (EditGuard) & $-0.0395$ & \textbf{0.1572} \\
        \textbf{WaterVIB (Ours)} & \textbf{0.0077} & \textbf{0.0306} \\
        \bottomrule
    \end{tabular}
    \end{small}
\end{table}

\textit{Analysis of Energy Disparity.} 
A seemingly negligible cosine similarity in high-dimensional spaces can be destructive due to the significant energy gap between the signal and the attack. In our experiments, the watermark is imperceptible ($\text{PSNR}_{wm} \approx 40$ dB), while the AIGC purification introduces substantial distortions ($\text{PSNR}_{atk} \approx 28$ dB). This $\approx 12$ dB difference implies that the AIGC noise vector is significantly stronger in magnitude than the watermark signal:
\begin{equation}
    \eta = \frac{||\mathbf{s}_{atk}||}{||\mathbf{s}_{wm}||} \approx 10^{\frac{40 - 28}{20}} \approx 3.98
\end{equation}
We define the \textit{Effective Projection} as the ratio of the destructive interference to the watermark strength: $E_{proj} = \eta \cdot |\cos \theta|$.
For the Baseline, the attack noise exerts a projection force equivalent to $\mathbf{15.7\%}$ ($3.98 \times 0.0395$) of the watermark's magnitude, effectively overwriting the signal. In contrast, WaterVIB achieves near-perfect orthogonality ($\cos \theta \approx 0.007$), suppressing the effective interference to a negligible $\mathbf{3.0\%}$, thus ensuring robustness despite the overwhelming energy of the generative process.

\textbf{Experiment II: Mechanism of Erasure via Negative Correlation.}
To further investigate \textit{how} generative purification destroys the watermark, we analyzed the statistical relationship between the generated AIGC distortion ($\mathbf{s}_{atk}$) and the original cover image content ($\mathbf{x}$). We measured both the global Cosine Similarity and the pixel-wise Pearson Correlation Coefficient (PCC) on the same 500-sample subset.

\begin{table}[h]
    \centering
    \caption{\textbf{Correlation Analysis: AIGC Distortion vs. Cover Image.} Unlike random noise (Gaussian) which is statistically orthogonal to the image content, AIGC distortion exhibits a significant negative correlation, indicating an active subtraction of image features.}
    \label{tab:correlation_analysis}
    \begin{small}
    \begin{tabular}{l|cc}
        \toprule
        \textbf{Noise Type} & \textbf{Cosine Sim.} ($\mathbf{s} \cdot \mathbf{x}$) & \textbf{Pearson Corr.} (PCC) \\
        \midrule
        Random Gaussian & $\approx 0.000$ & $\approx 0.000$ \\
        \textbf{AIGC (SDXL 1.0)} & \textbf{-0.1918} & \textbf{-0.3399} \\
        \bottomrule
    \end{tabular}
    \end{small}
\end{table}

\textit{Analysis of Content Suppression.}
The negative correlations reported in Table \ref{tab:correlation_analysis} reveal the fundamental mechanism of the attack.
A correlation of approximately zero (as seen in random noise) implies that the distortion is independent of the content. However, the significant negative values ($\text{PCC} \approx -0.34$) observed for AIGC indicate that the generated distortion $\mathbf{s}_{atk}$ is \textit{structurally opposed} to the cover image $\mathbf{x}$.
Mathematically, the purified image is formed as $\mathbf{x}_{pure} = \mathbf{x} + \mathbf{s}_{atk}$. A negative projection ($\mathbf{s}_{atk} \cdot \mathbf{x} < 0$) implies that the addition of $\mathbf{s}_{atk}$ reduces the magnitude of the original images:
\begin{equation}
    ||\mathbf{x} + \mathbf{s}_{atk}|| < ||\mathbf{x}|| \quad \text{(along aligned dimensions)}
\end{equation}
This confirms that the generative model acts as a \textbf{Content-Adaptive Eraser}. It perceives high-frequency details (which harbor the watermark) as "perceptual noise" or "energetic costs" and generates a counter-signal to smooth or suppress them. This active erasure explains why texture-entangled watermarks are obliterated even when the visual change appears minimal.

\subsection{Visualization of Watermark Residual Distribution}
\label{appendix:qualitative_residual}

\begin{figure}[h]
    \centering
    \includegraphics[width=1.0\linewidth]{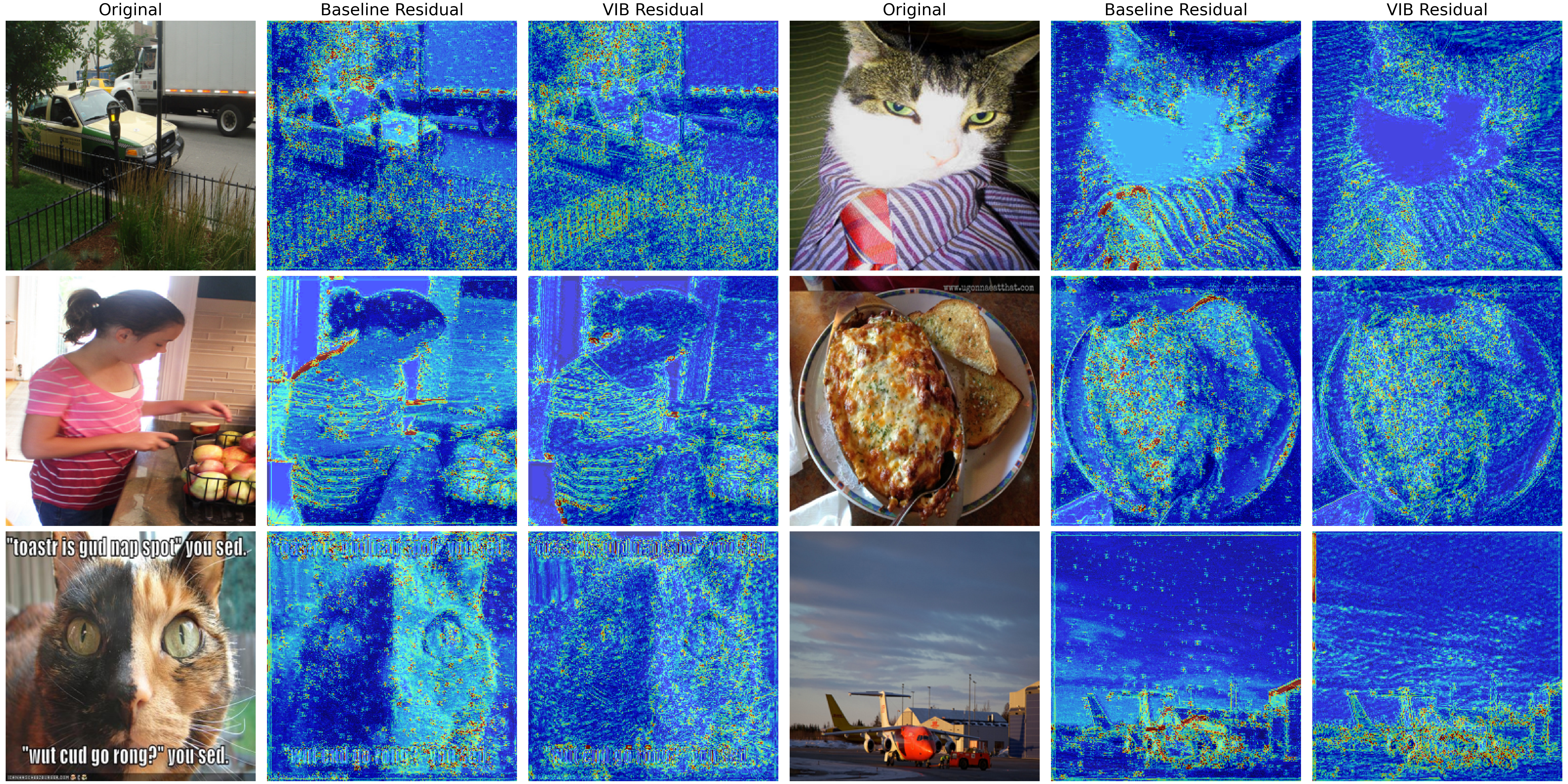}
    \caption{\textbf{Qualitative Comparison of Watermark Residuals.} We visualize the absolute difference $|x_{wm} - x|$ (amplified for visibility) between the watermarked and original images. The \textbf{Baseline} (HiDDeN) residuals are heavily entangled with high-frequency image textures. In contrast, \textbf{WaterVIB} produces a more dispersed noise pattern, indicating that the Information Bottleneck constraint successfully decouples the watermark signal from fragile cover details.}
    \label{fig:residual_comparison}
\end{figure}

To provide an intuitive understanding of how the Information Bottleneck principle alters the embedding strategy, we visualize the spatial distribution of the watermark signal. Figure~\ref{fig:residual_comparison} displays the absolute residual maps $|x_{wm} - x|$ for both the Baseline (HiDDeN) and our WaterVIB model across diverse cover images.

\noindent\textbf{Texture Entanglement and Spatial Clustering in Baseline.} 
As observed in the ``Baseline Residual'' column, the standard encoder exhibits a strong \textit{structural dependency} on the cover image, with watermark energy heavily concentrated and clustered on specific high-frequency object contours and sharp edges. This localization corroborates our analysis in Section 3.1: the baseline model implicitly learns to hide information within the most complex textures to satisfy invisibility constraints. However, this creates a critical vulnerability, as generative purification models specifically target and rewrite these high-gradient texture regions during the manifold projection process, effectively erasing the watermark along with the original texture.

\noindent\textbf{Uniform Texture Diffusion in WaterVIB.} 
In contrast, the ``VIB Residual'' column demonstrates that WaterVIB induces a significantly more \textit{uniform diffusion} of the signal across the textured domains of the image. Instead of overfitting the watermark to a limited set of sharp edges, the VIB-constrained model spreads the signal intensity across a broader and more diverse range of structural features. By enforcing the Information Bottleneck constraint to minimize $I(Z;X)$, the encoder is compelled to abandon its reliance on specific, fragile texture details. While the signal remains predominantly within the textured regions to maintain imperceptibility, its distribution is no longer tied to the narrow set of pixels most susceptible to generative reconstruction. This expanded and more uniform coverage within the textured manifold ensures that the watermark information remains resilient even when generative models perform localized editing or semantic rewriting on the image's high-frequency details.

% ==========================================
% Appendix F: Defense against Re-Embedding
% ==========================================

\section{Defense against Re-Embedding Attacks}
\label{app:re_embedding}

In real-world API deployment scenarios, a potential threat arises from \textit{Re-Embedding Attacks} (also known as Multi-Watermarking), where an adversary attempts to overwrite or confuse the existing watermark by embedding a new message into an already watermarked image.

\subsection{Defense Mechanism: Pre-Embedding Detection}

To mitigate this threat, we implement a \textbf{detection-based defense mechanism}. The system enforces a strict policy: \textit{no new embedding is permitted if a watermark is already detected.} This requires the decoder to function not only as a message extractor but also as a robust binary detector (Watermarked vs. Clean).

We leverage the output logits of the decoder as a confidence metric. Specifically, we define the \textbf{Average Logits (AL)} score:
\begin{equation}
    AL = \frac{1}{L} \sum_{i=1}^{L} |l_i|
    \label{eq:avg_logits}
\end{equation}
where $l_i$ represents the raw logit output for the $i$-th bit of the message. Due to the sigmoid activation in the final layer, higher absolute logits $|l_i|$ correspond to higher confidence predictions (approaching 0 or 1).

\subsection{The VIB Advantage in Detection}

The WaterVIB framework inherently enhances detectability. The Information Bottleneck objective ($\mathcal{L}_{IB}$) constrains the latent representation to capture only the \textit{Minimal Sufficient Statistics} of the message. This optimization pressure forces the decoder to be extremely decisive, pushing the output distribution towards deterministic states (high confidence) for watermarked images, while clean images (which lack the specific watermark structure) produce low-confidence, high-entropy outputs.

This distributional separation effectively widens the gap between the two classes, making the detection threshold robust even under noise.

\subsection{Empirical Evaluation}

We evaluated this defense on a held-out set of 1,500 images: 500 clean images, 500 watermarked images embedded with random messages, and 500 watermarked images subjected to Gaussian noise ($\sigma=0.05$). We compare the detection performance of our \textbf{HiDDeN-VIB} against the standard \textbf{HiDDeN} baseline.

\paragraph{Metrics.} We report the False Positive Rate (\textbf{FP}) on clean images, False Negative Rate (\textbf{FN}) on watermarked images, and the \textbf{FN} under noise. Additionally, we measure the Kullback-Leibler Divergence (\textbf{KL-Div}) between the logit distributions of clean and watermarked samples to quantify separability.

\paragraph{Results.} As shown in Table~\ref{tab:re_embedding_results}, both models achieve perfect separation (0\% error) on clean and standard watermarked images. However, under noise attacks, the baseline's detection capability degrades significantly (FN rises to 38.50\%). In contrast, WaterVIB maintains robust detection with an FN of only \textbf{1.98\%}. The higher KL-Divergence (\textbf{3.84} vs. 3.33) confirms that WaterVIB learns a more distinct and robust watermark manifold, effectively preventing re-embedding attacks even when the image has been degraded.

\begin{table}[h]
    \centering
    \caption{\textbf{Re-Embedding Defense Performance.} Comparison of detection robustness. The VIB constraint significantly improves detection under noise, preventing unauthorized re-embedding.}
    \label{tab:re_embedding_results}
    \vspace{5pt} % 适当调整间距
    \resizebox{0.9\linewidth}{!}{ % 如果表格太宽，自动缩放
    \begin{tabular}{l c c c c}
        \toprule
        \textbf{Model} & \textbf{FP} (Clean) & \textbf{FN} (Watermarked) & \textbf{FN} (WM + Noised) & \textbf{KL-Div} $\uparrow$ \\
        \midrule
        HiDDeN (Baseline) & 0.00\% & 0.00\% & 38.50\% & 3.33 \\
        \textbf{HiDDeN-VIB (Ours)} & \textbf{0.00\%} & \textbf{0.00\%} & \textbf{1.98\%} & \textbf{3.84} \\
        \bottomrule
    \end{tabular}
    }
\end{table}

\end{document}